\documentclass[sn-mathphys-num]{sn-jnl}

\usepackage{graphicx}%
\usepackage{epsfig}
\usepackage{multirow}%
\usepackage{amsmath,amssymb,amsfonts}%
\usepackage{siunitx,makecell,booktabs}
\usepackage{amsthm}%
\usepackage{mathrsfs}%
\usepackage[title]{appendix}%
\usepackage{xcolor}%
\usepackage{textcomp}%
\usepackage{manyfoot}%
\usepackage{booktabs}%
\usepackage{algorithm}%
\usepackage{algorithmicx}%
\usepackage{algpseudocode}%
\usepackage{listings}%
\usepackage{titlesec}
\usepackage{pifont}
\usepackage{subcaption} 
\usepackage{array}
\usepackage{tabularx}
\usepackage[numbers]{natbib}
\theoremstyle{thmstyleone}%
\theoremstyle{thmstyletwo}%

\theoremstyle{thmstylethree}%

\begin{document}

\title{DPSformer: A long-tail-aware model for improving heavy rainfall prediction}


\author[1, 2]{\fnm{Zenghui} \sur{Huang}}
\equalcont{These authors contributed equally to this work.}

\author[3, 4]{\fnm{Ting} \sur{Shu}}
\equalcont{These authors contributed equally to this work.}

\author*[5]{\fnm{Zhonglei} \sur{Wang}} 
\email{wangzl@xmu.edu.cn}

\author*[1, 2]{\fnm{Yang} \sur{Lu}} 
\email{luyang@xmu.edu.cn}

\author[1, 2]{\fnm{Yan} \sur{Yan}} 

\author[5]{\fnm{Wei} \sur{Zhong}} 

\author[1, 2]{\fnm{Hanzi} \sur{Wang}} 

\affil[1]{Key Laboratory of Multimedia Trusted Perception and Efficient Computing, Ministry of Education of China, Xiamen University, Xiamen, China}

\affil[2]{Fujian Key Laboratory of Sensing and Computing for Smart City, School of Informatics, Xiamen University, Xiamen, China}

\affil[3]{\orgdiv{School of Artificial Intelligence}, \orgname{Shenzhen University}, \orgaddress{\city{Shenzhen}, \postcode{518060}, \country{China}}}

\affil[4]{\orgdiv{National Engineering Laboratory for Big Data System Computing Technology}, \orgname{Shenzhen University}, \orgaddress{\city{Shenzhen}, \postcode{518060}, \country{China}}}

\affil[5]{\orgdiv{Wang Yanan Institute for Studies in Economics}, \orgname{Xiamen University}, \orgaddress{\city{Xiamen}, \postcode{361005}, \country{China}}}

\titleformat{\subsection}
  {\normalfont\large\bfseries}
  {}{0em}{}


\abstract{Accurate and timely forecasting of heavy rainfall remains a critical challenge for modern society. Precipitation exhibits a highly imbalanced distribution: most observations record no or light rain, while heavy rainfall events are rare. Such an imbalanced distribution obstructs deep learning models from effectively predicting heavy rainfall events. To address this challenge, we treat rainfall forecasting explicitly as a long‑tailed learning problem, identifying the insufficient representation of heavy rainfall events as the primary barrier to forecasting accuracy. Therefore, we introduce DPSformer, a long‑tail‑aware model that enriches representation of heavy rainfall events through a high‑resolution branch. For heavy rainfall events $\geq$ 50~mm/6~h, DPSformer lifts the Critical Success Index (CSI) of a baseline Numerical Weather Prediction (NWP) model from 0.012 to 0.067. For the top~1\% coverage of heavy rainfall events, its Fraction Skill Score (FSS) exceeds 0.45, surpassing existing methods. Our work establishes an effective long-tailed paradigm for heavy rainfall prediction, offering a practical tool to enhance early warning systems and mitigate the societal impacts of extreme weather events.}




\maketitle

\section{Introduction}\label{sec1}
Accurate forecasting of heavy rainfall is critical for public safety, disaster prevention, and economic operations, serving as a cornerstone of modern society~\cite{Seneviratne2021}. Despite this urgent need, the complex and heterogeneous spatiotemporal patterns of heavy rainfall events continue to challenge the capabilities of existing precipitation forecasting models~\cite{olivetti2024advances}. Among these, Numerical Weather Prediction (NWP) models have played a critical role in short-term precipitation forecasting by utilizing complex physical equations and high-performance computing to simulate atmospheric processes~\cite{bauer2015quiet, ref1,ref2,ref3}. Yet, the predictive skill of NWP models is markedly reduced when confronted with highly nonlinear and localized heavy rainfall~\cite{chen2015precipitation}. This deficiency manifests as critical forecast failures, such as spatial displacement errors and underestimation of the intensity of heavy rainfall~\cite{yuval2020stable}. Such failures are largely attributable to fundamental limitations within the NWP framework, including the propagation of initial condition uncertainties and poorly constrained convective parameterization~\cite{Zhang2014comparative, buizza2005comparison}. In parallel, the rise of deep learning has brought about a new generation of purely data-driven, large-scale weather models~\cite{price2025probabilistic, chen2023machine, sonderby2020metnet}. Trained on massive reanalysis datasets, these models achieve accuracy comparable to or even surpassing that of state-of-the-art NWP models, while offering several orders of magnitude greater computational efficiency~\cite{espeholt2022deep, pathak2022fourcastnet, bi2023accurate, das2024hybrid}. Nevertheless, these models are primarily developed for general forecasting tasks and lack dedicated optimization for low-frequency, high-impact heavy rainfall events, which consequently impairs their performance in predicting heavy rainfall~\cite{ji2024leadsee}. Both NWP and deep learning models, despite their respective strengths, struggle to reliably predict the most severe rainfall events~\cite{verma2023deep}. This highlights the urgent need for a new forecasting paradigm better equipped to anticipate the risks posed by heavy rainfall~\cite{CampsValls2025}.

\begin{figure}
  \centering
  \includegraphics[width=1\textwidth]{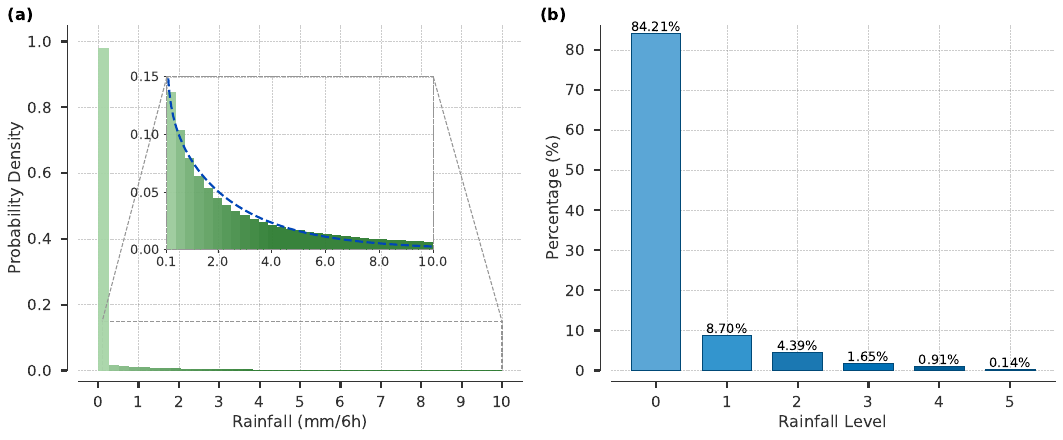} 
  \caption{\textbf{Rainfall distribution in the training dataset.} (a) Probability density distribution of rainfall (mm/6~h) in the training dataset, with a fit obtained using a gamma distribution. (b) Percentage distribution of rainfall classified into six levels based on predefined thresholds (0.1, 3, 10, 20, and 50~mm/6~h).} 
  \label{img1} 
\end{figure}

The fundamental challenge in heavy rainfall prediction is rooted in the statistical nature of precipitation itself: a characteristic long-tailed distribution~\cite{zhang2023deep, shi2025deep}. As shown in Fig.~\ref{img1}, the vast majority of observations correspond to no or light rain, while heavy rainfall events occur only rarely. This statistical rarity hinders the robust validation and tuning of physical parameterizations in NWP models~\cite{Reichstein2019}. For data-driven deep learning models, this pronounced data imbalance leads them to be predominantly influenced by frequent, low-intensity (head-class) rainfall during training~\cite{olivetti2024advances}, resulting in a persistent bias toward weaker rainfall predictions and poor performance for rare, high-intensity (tail-class) events~\cite{ji2024leadsee}. While long-tailed learning has emerged as a vibrant research area in deep learning for tackling such imbalanced data, its application in meteorology remains nascent. Long-tailed learning theories posit that failures in predicting rare events stem from two entangled factors: inadequate feature representation and skewed classifier decision boundaries~\cite{yang2022survey}. Prevailing methods have often attempted to address this challenge holistically, typically by modifying the loss function with the implicit hope of concurrently resolving both issues~\cite{tian2024post, hess2022deep}. However, this one-size-fits-all strategy proves insufficient in the complex domain of precipitation forecasting~\cite{you2023study, tang2023postrainbench}. Given that meteorological data exhibits a complexity far surpassing that of natural images~\cite{sun2025data}, simply adjusting the decision boundary yields limited effectiveness~\cite{kang2019decoupling}. The fundamental challenge stems from the difficulty in learning robust and discriminative feature representation for rare heavy rainfall events~\cite{li2025swinnowcast}. This highlights that the main bottleneck in long-tailed rainfall forecasting is foundational representation learning, requiring a strategy that directly aims at enhancing representation quality.

In this study, we introduce a long-tailed learning perspective for heavy rainfall prediction. Our core strategy moves beyond conventional methods by directly targeting the problem of representation deficiency. To this end, we propose a Dual-Path Spatial feature enhancement SegFormer (DPSformer) architecture to post-process NWP precipitation outputs, comprising a standard backbone and a dedicated high-resolution spatial branch. The spatial branch is intentionally designed to extract fine-grained spatial information associated with heavy rainfall, thereby alleviating the feature underrepresentation induced by the long-tailed distribution. Through extensive experiments, we show that our framework achieves substantial improvements in correcting NWP errors for heavy rainfall events. Notably, for heavy rainfall events $\geq 50$~mm/6~h, our framework improves the Critical Success Index (CSI) by over 450\% compared to the original NWP, raising the score from 0.012 to 0.067. For the top~1\% coverage of heavy rainfall events, its Fraction Skill Score (FSS) exceeds 0.45, outperforming existing models. Our work establishes a new paradigm for tackling heavy rainfall prediction from the perspective of long-tailed learning.

\begin{figure}
  \centering
  \includegraphics[width=1\textwidth]{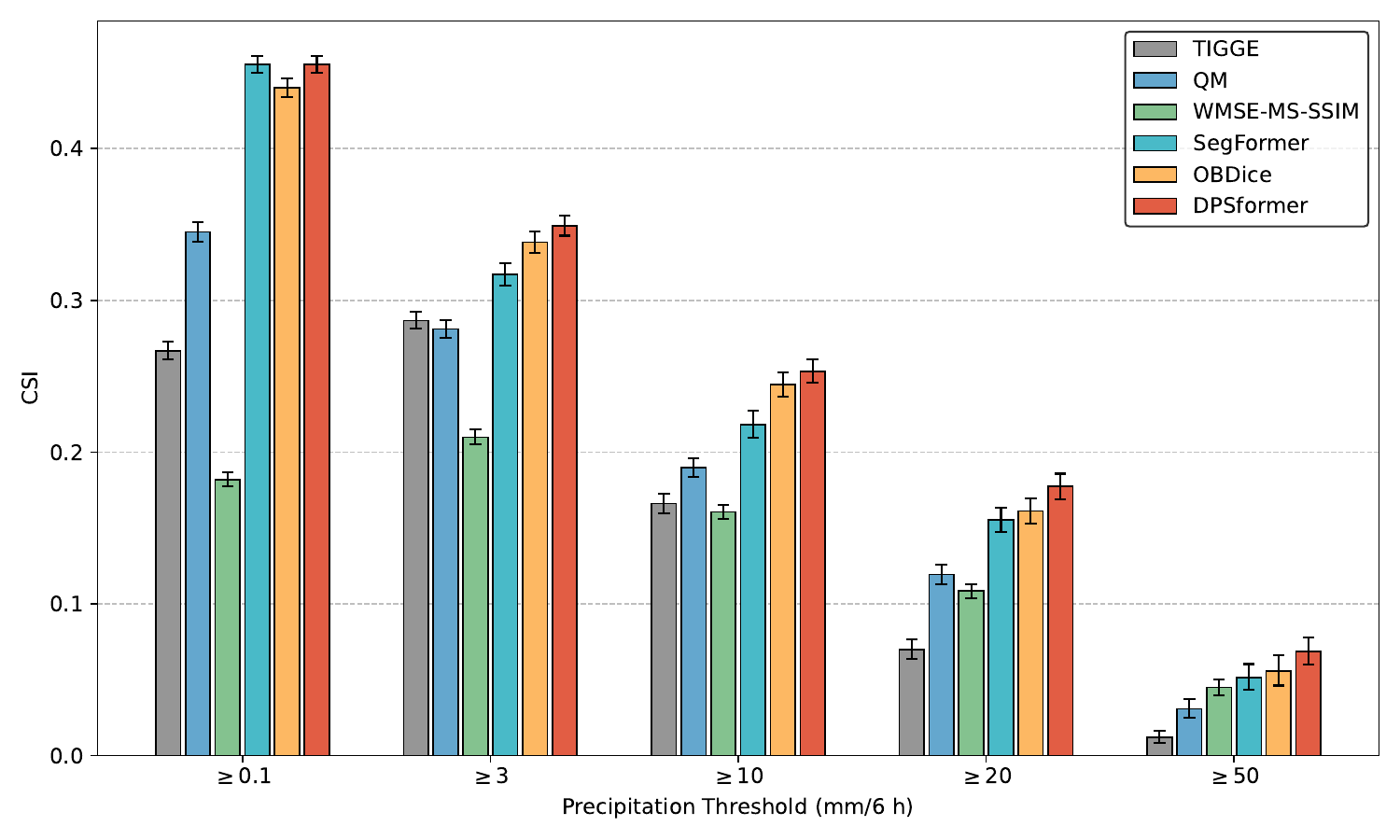} 
  \caption{\textbf{Performance comparison of baseline methods under long-tailed rainfall conditions.}  
CSI scores of DPSformer and five baselines evaluated at five increasing 6-h accumulated rainfall thresholds. Error bars denote 95\% confidence intervals, estimated via bootstrap (n=1000).
} 
  \label{img2} 
\end{figure}  

\section{Results}\label{sec2}

\subsection*{Comprehensive performance analysis}\label{subsec21}
Our proposed DPSformer improves the prediction of heavy rainfall events, demonstrating an advantage in forecasting the most widespread and high-impact rainfall events where traditional models typically struggle. To substantiate this claim, we compare DPSformer with a series of representative post-processing baseline models, including the raw THORPEX Interactive Grand Global Ensemble (TIGGE) forecast~\cite{swinbank2016tigge, bougeault2010thorpex}, traditional statistical methods quantile mapping (QM)~\cite{maraun2010precipitation}, a standard deep learning model (SegFormer~\cite{xie2021segformer}), and deep learning NWP post-processing methods (WMSE-MS-SSIM~\cite{hess2022deep} and OBDice~\cite{you2023study}). All models are trained and evaluated on the same subset of the European Centre for Medium-Range Weather Forecasts (ECMWF) TIGGE forecast archive, with a spatial resolution of 0.25° × 0.25°. The prediction target is 6-hourly accumulated rainfall. The comparative analysis relies on seven standard categorical metrics: the CSI and Equitable Threat Score (ETS) to gauge overall forecast skill; the Probability of Detection (POD), False Alarm Ratio (FAR), and Miss Alarm Ratio (MAR) to assess event detection accuracy; the F1-score for the precision-recall balance; and the Symmetric Extremal Dependence Index (SEDI)~\cite{ferro2011extremal}, which is specifically used to evaluate extreme events. This set of metrics provides a robust basis for assessing the relative effectiveness of the models. All comparisons are performed at five rainfall thresholds: 0.1, 3, 10, 20, and 50~mm/6~h, with the latter two corresponding to heavy rainfall events~\cite{you2023study}. Unless otherwise stated, all results are the mean of experiments performed with five different random seeds. More detailed settings and reproducibility resources are provided in Supplementary Table~\ref{tab:all_hyperparams}.

\begin{table}[t]
\centering
\caption{
\textbf{Performance comparison of baseline methods for heavy rainfall events ($\geq 50$~mm/6~h).}
$\Delta$SEDI denotes the improvement in SEDI relative to the TIGGE baseline. 
}
\label{tab:50mm6h}
\begin{tabular}{lccccccc}
\toprule
\multirow{1}{*}{\textbf{Method}} & CSI\,\raisebox{+0.25ex}{$\uparrow$} & ETS\,\raisebox{+0.25ex}{$\uparrow$} & POD\,\raisebox{+0.25ex}{$\uparrow$} & MAR\,\raisebox{+0.25ex}{$\downarrow$} & FAR\,\raisebox{+0.25ex}{$\downarrow$} & F1\,\raisebox{+0.25ex}{$\uparrow$} & $\Delta$SEDI\,\raisebox{+0.25ex}{$\uparrow$} \\
\midrule
TIGGE      & 0.012 & 0.012 & 0.013 & 0.987 & \textbf{0.881} & 0.024   & -- \\
QM & 0.031 & 0.030 & 0.050 & 0.950 & 0.926 & 0.060 & 0.003  \\
SegFormer  & 0.050 & 0.049 & 0.126 & 0.873 & 0.920 & 0.096 & 0.049 \\
WMSE-MS-SSIM  & 0.048 & 0.047 & \underline{0.170} & \underline{0.830} & 0.932 & 0.092 & \underline{0.066} \\
OBDice     & \underline{0.056} & \underline{0.055} & 0.108 & 0.892 & \underline{0.895} & \underline{0.106} & 0.053 \\
DPSformer & \textbf{0.067} & \textbf{0.066} & \textbf{0.220} & \textbf{0.780} & 0.911 & \textbf{0.126} & \textbf{0.117} \\
\bottomrule
\end{tabular}

\begin{tablenotes}[flushleft]
  \footnotesize
  \item \raisebox{+0.25ex}{$\uparrow$} Indicates that higher values are better for the corresponding metric.
  \item \raisebox{+0.25ex}{$\downarrow$} Indicates that lower values are better for the corresponding metric.
  \item The best result in each column is shown in \textbf{bold}, and the second-best is \underline{underlined}.
  \end{tablenotes}

\end{table}

To quantitatively assess the advantages of DPSformer, we compare its performance with mainstream post-processing baseline models at the 50~mm/6~h threshold (Table~\ref{tab:50mm6h}). Figure~\ref{img2} further shows DPSformer's consistently superior CSI across all thresholds. DPSformer achieves more noticeable improvements at higher rainfall thresholds, where accurate detection is both critical and challenging. For heavy rainfall events $\geq$ 50~mm/6~h, DPSformer achieves a CSI of 0.067 and a POD of 0.220, over five times and fifteen times higher than the raw NWP forecasts (TIGGE's CSI of 0.012 and POD of 0.013), respectively. It also obtains the highest F1-score (0.126), reflecting its capacity to effectively balance detection rates with precision. Furthermore, DPSformer demonstrates stronger discrimination ability for extreme events, as evidenced by its leading SEDI improvement of 0.117, surpassing other methods such as OBDice (0.053) and SegFormer (0.049). The SEDI metric is particularly meaningful for evaluating rare event forecasting as it remains equitable even under class imbalance. This advantage in SEDI, coupled with its balance between a higher POD and relatively lower MAR (0.780) and FAR (0.911) compared to other models, underscores its capability in identifying and localizing heavy rainfall events. High rates of missed detections and false alarms are an inherent challenge in extreme event forecasting. Despite this, DPSformer achieves a more effective trade--off than competing methods. These results indicate that DPSformer provides more accurate and reliable predictions for rare, high-intensity rainfall events under long-tailed distributions. The expanded metric suite for both 20~mm/6~h and 50~mm/6~h thresholds is listed in Supplementary Table~\ref{tab:detailed_results}. Ablation experiments (Supplementary Tables~\ref{tab:ablation_50mm} and \ref{tab:Parameter size}) corroborate that DPSformer’s performance gains arise from architectural innovations rather than parameter scaling. In addition, feature quality comparisons (Supplementary Table~\ref{tab:feature_metrics_appendix}) show that DPSformer improves the separability of learned representation, underscoring the effectiveness of enhancing feature quality for rare heavy rainfall events.

\begin{figure}
  \centering
  \includegraphics[width=1.0\textwidth]{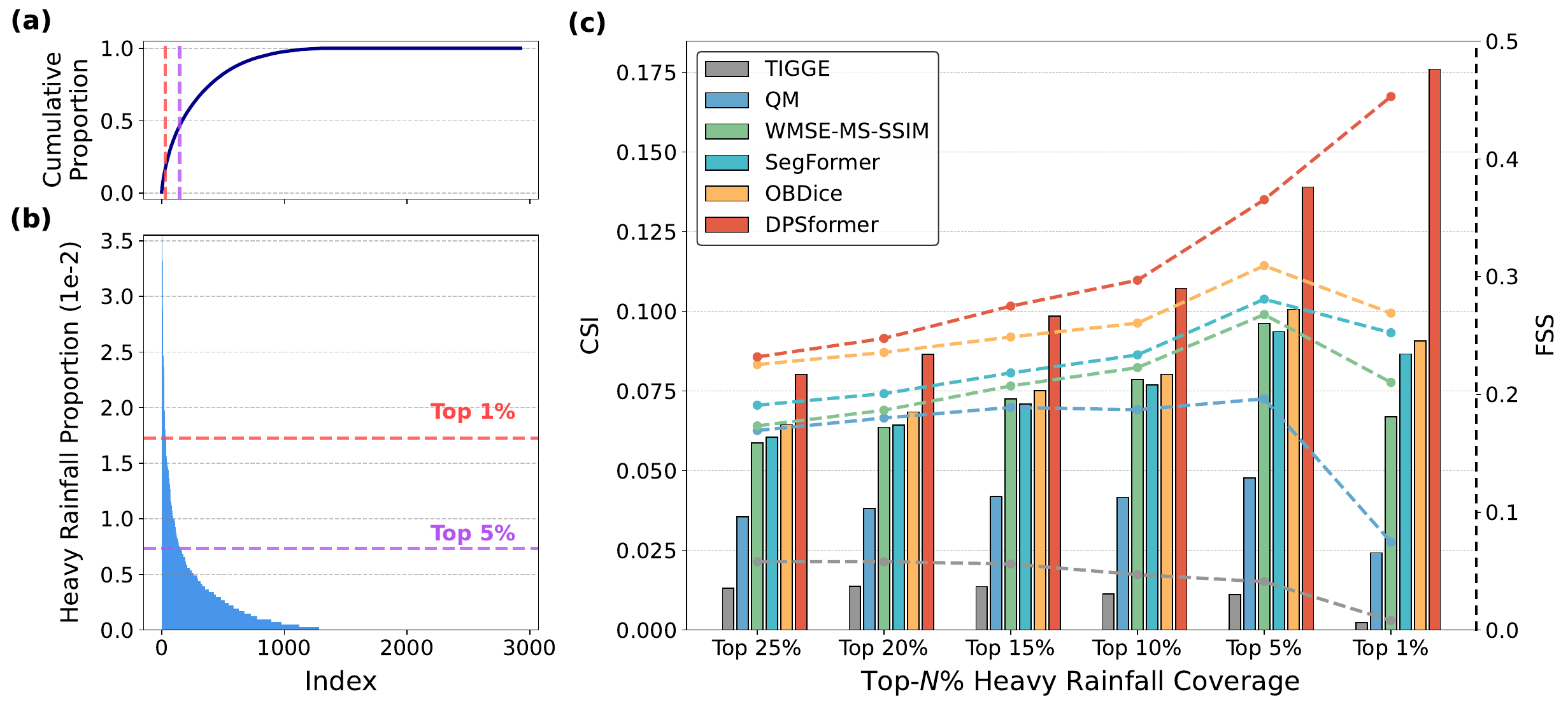} 
  \caption{\textbf{Performance comparison of baseline methods for high-coverage heavy rainfall events.} (a) Cumulative distribution of the occurrence proportion of heavy rainfall ($\geq $50 mm / 6 h) in the test dataset, with lines marking the top~5\% and top~1\% percentiles. (b) Probability distribution of the high-coverage heavy rainfall events across all test dataset. (c) CSI and FSS performance of different methods for high-coverage heavy rainfall events. Evaluations are based on rainfall maps ranked by descending order of heavy rainfall coverage percentage (from top~25\% to top~1\%). Bar heights represent CSI values (left y-axis), while the overlaid lines indicate FSS scores (right y-axis) for the respective models. FSS is calculated using a neighborhood size of 5 grid points.} 
  \label{img3} 
\end{figure}  

To further probe DPSformer's capabilities under the most challenging conditions, we conducted a targeted evaluation that focused specifically on test cases with the highest coverage of heavy rainfall. We sorted the test dataset by its heavy rainfall coverage and evaluated model performance within the top~N\% of these cases, as illustrated in Fig.~\ref{img3}(a) and~\ref{img3}(b), with the results shown in Fig.~\ref{img3}(c). This analysis highlights a key strength of our proposed method. As the evaluation shifts towards more extreme scenarios (from the top~25\% to the top~1\%), the performance gap between DPSformer and all other methods becomes increasingly evident. In the top~1\% of the most widespread rainfall cases, DPSformer achieves a CSI of approximately 0.18, which is double that of the second-best method, OBDice (CSI = 0.09), and more than 50 times higher than that of the TIGGE baseline (CSI = 0.002). A similar trend is observed for the FSS, where DPSformer's score exceeds 0.4, indicating superior performance in capturing the spatial structure of the rainfall fields, while other methods fail to produce coherent forecasts (FSS $<$ 0.3). This result demonstrates that DPSformer does not just offer a marginal average improvement, but shows clear advantages in predicting the most severe and high-impact weather events, where conventional methods remain limited. This robust performance at the tail-end of the distribution provides compelling evidence that DPSformer effectively addresses the long-tailed challenge in rainfall forecasting, making it a more reliable tool for operational high-impact weather prediction.

\begin{figure}
  \centering
  \includegraphics[width=1.0\textwidth]{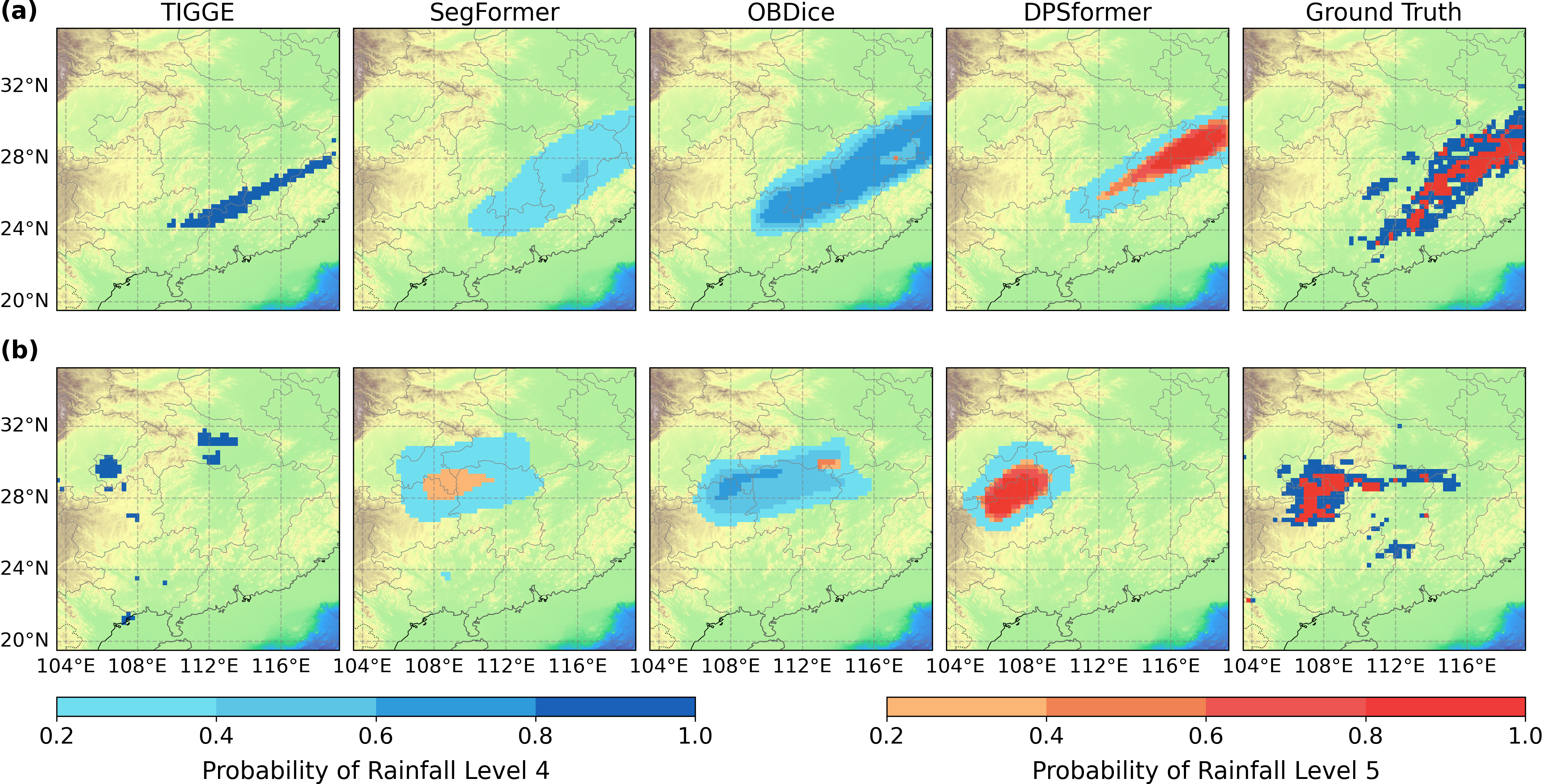} 
  \caption{\textbf{Warm-sector mesoscale convective system (MCS) heavy rainfall events.} This figure compares the ability of TIGGE and three deep learning methods (SegFormer, OBDice and DPSformer) to localize heavy rainfall events. The comparison is shown for two heavy rainfall categories: Level 4 ([20, 50)~mm/6~h) and Level 5 ($\geq$50~mm/6~h). While the ground truth and the TIGGE output are represented as binary occurrence maps (presence/absence), the deep learning methods generate probability fields. (a) shows the case for 2012.03.04 18:00-24:00 UTC; (b) shows the case for 2012.05.11 18:00-24:00 UTC.} 
  \label{img4} 
\end{figure}  

To highlight DPSformer's capability in capturing the spatial distribution of heavy rainfall events, we present its rainfall probability field alongside the ground truth and raw TIGGE forecasts in Fig.~\ref{img4}. DPSformer allocates elevated probabilities to the majority of areas exhibiting actual heavy rainfall events, thereby accurately reflecting the geographic spread of intense rainfall. In case (a), DPSformer precisely captures an elongated heavy rainfall belt stretching across southeastern coastal and inland regions, aligning closely with the fragmented yet slender pattern in the ground truth. By contrast, TIGGE, although it marginally detects this rain belt, severely underestimates the intensity of the heavy rainfall. In case (b), DPSformer robustly identifies the rainfall core in a central mountainous area, with its high-confidence regions better encompassing the clustered extreme events observed in the ground truth, while TIGGE is largely ineffective. The strong spatial overlap between DPSformer predictions and the ground truth demonstrates not only improved detection of high-impact rainfall but also a high level of confidence in its spatial localization, an essential property for early warning and decision support applications.

To further quantify the uncertainty of DPSformer, besides the bootstrap confidence intervals reported for CSI, we quantify instance‑level uncertainty with 95\% conformal prediction, specifically a Mondrian‑Conformal Classification (MCC) scheme~\cite{sun2017applying}. Across the test dataset, DPSformer achieves a pixel‑wise coverage (PICP) of 94.6\%, close to the nominal 95\%, indicating well‑calibrated prediction sets. The average prediction‑set size is 2.8 out of 6 possible rainfall classes, demonstrating that high coverage is obtained with a compact decision set and thus efficient discrimination. The slight under‑coverage remains acceptable for operational use, supporting the reliability of our uncertainty quantification.

\subsection{Tropical cyclone rainfall forecasting results}\label{subsec22}
 
\begin{figure}
  \centering
  \includegraphics[width=1\textwidth]{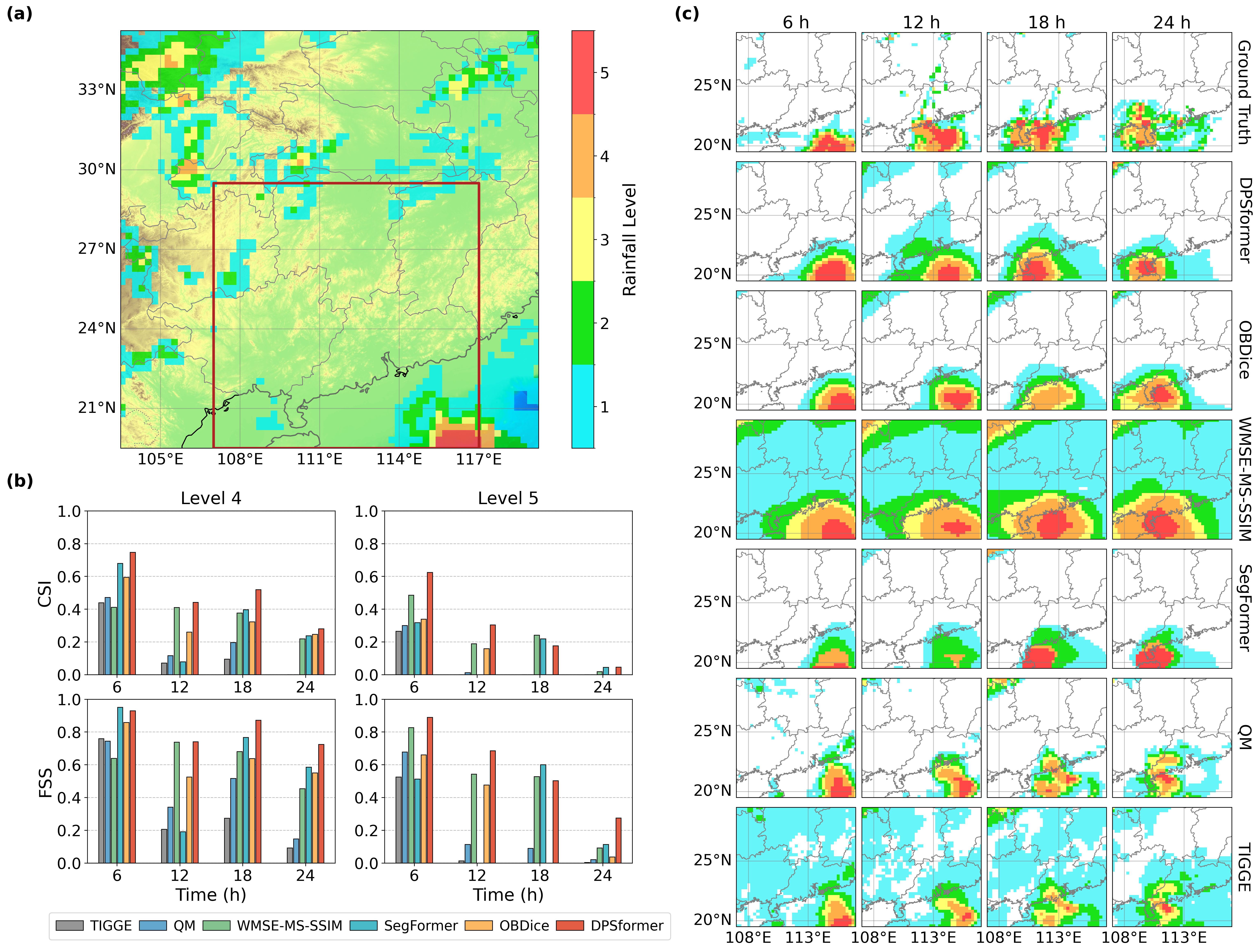} 
  \caption{\textbf{Tropical cyclone heavy rainfall (Tropical Storm Doksuri).} 
Event date: 2012-06-29 (UTC). 
(a) Study region used for evaluation, outlined in red on the rainfall map, together with the 6-hour accumulated rainfall from 18--24~h UTC on the previous day, representing the initial rainfall state before the case day. 
(b) Performance comparison of DPSformer and five baselines in terms of CSI and FSS at heavy rainfall Level 4 ([20, 50)~mm/6~h) and Level 5 ($\geq$50~mm/6~h), evaluated for four consecutive 6-hour periods within the case day. FSS is calculated using a neighborhood size of 5 grid points.
(c) Spatial distribution of 6-hourly accumulated rainfall over four consecutive periods within that day (00--06~h, 06--12~h, 12--18~h, and 18--24~h UTC) from ground truth, DPSformer, and the five baselines.} 
  \label{img5} 
\end{figure}  

To provide a more intuitive and qualitative assessment of DPSformer's performance, we present a case study analyzing the temporal evolution of a high-impact weather event: Tropical Storm Doksuri on June 29, 2012. Figure~\ref{img5} visualizes the 6-hourly accumulated rainfall from the TIGGE, our DPSformer's correction, and the ground truth over four consecutive periods within that day (00-06 h, 06-12 h, 12-18 h, and 18-24~h UTC). This analysis showcases DPSformer's ability to provide consistent and accurate guidance as the event evolves. This case illustrates the characteristic errors of the TIGGE forecast and the substantial improvements offered by DPSformer. In the initial 00-06 h period, TIGGE correctly predicts the presence of rainfall but fails to resolve the intense, organized rainband, underestimating its coverage and leading to a modest FSS of 0.526 for heavy rainfall ($\geq$ 50~mm/6~h). In contrast, DPSformer not only accurately delineates the shape and location of the rainband but also enhances its intensity to match the ground truth, achieving a near-perfect FSS of 0.889. As the day progresses, TIGGE's inability to capture the storm's dynamics becomes more pronounced. During the 06-12 h and 12-18 h periods, TIGGE fails to capture the northwestward propagation of the storm's core rainfall structure. Its forecasts depict only weak, scattered rainfall, severely underestimating the heavy rainfall and causing the FSS to collapse to near-zero values (0.013 and 0.000, respectively). DPSformer, however, demonstrates its remarkable ability to correct these severe spatial and intensity biases in each period. It successfully tracks the evolution of the rainband as it moves inland, maintaining high fidelity to the ground truth in both location and structure, as reflected by its consistently high FSS scores of 0.686 and 0.502. In the final period of the day (18-24 h), TIGGE exhibits a significant displacement error, completely missing the storm's residual rainband. In comparison, DPSformer still successfully captures its primary structure. This continuous, period-by-period analysis provides visual evidence that DPSformer can effectively rectify the systematic deficiencies of the NWP model throughout the life cycle of a structured, high-impact convective system, thereby offering potential contribution for operational short-term forecasting.

\begin{figure}
  \centering
  \includegraphics[width=1.0\textwidth]{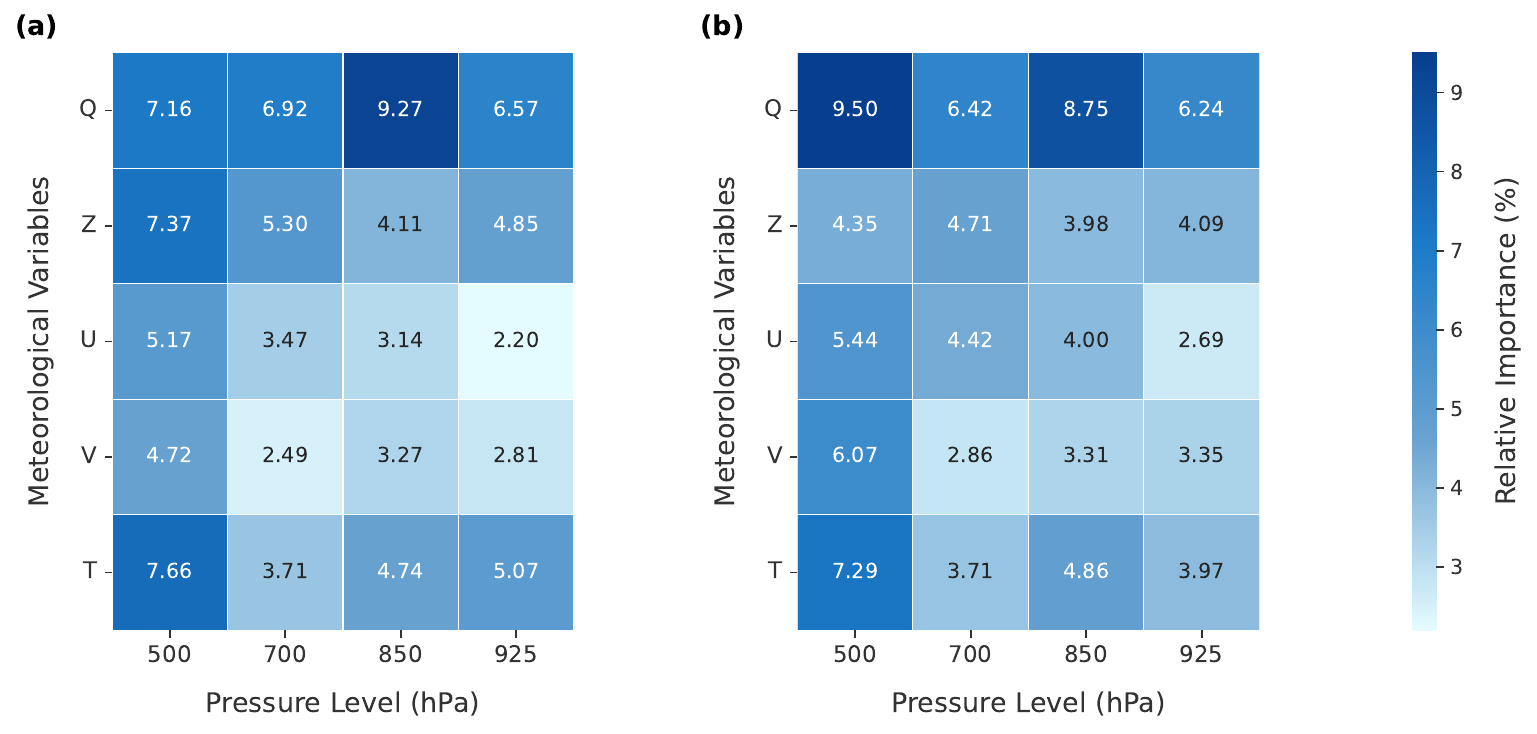} 
  \caption{\textbf{Vertical attribution profiles of key meteorological variables at different pressure levels.} The values represent relative importance obtained by normalizing the IG scores across all pressure level variables. (a) All pixels within the evaluation domain. (b) heavy rainfall pixels only. Rows correspond to five representative atmospheric variables—specific humidity (Q), geopotential height (Z), U component of wind (U), V component of wind (V), and temperature (T)—and columns correspond to four pressure levels (500, 700, 850, and 925~hPa).} 
  \label{img444545} 
\end{figure} 

\subsection{Analyzing effects of meteorological variables}\label{subsec23} 
To evaluate DPSformer's ability to capture the physical mechanisms underlying rainfall formation, we performed an attribution analysis on all 27 input meteorological variables (see Method) using the Integrated Gradients (IG) method~\cite{sundararajan2017axiomatic}. Our attribution analysis demonstrates that DPSformer has learned an adaptive and physically coherent strategy, fundamentally altering its focus based on rainfall intensity. This is most evident in the vertical atmospheric profiles (Fig.~\ref{img444545}). As shown in Fig.~\ref{img444545}(a), for general rainfall, DPSformer balances upper-level dynamics (geopotential height (Z) at 500 hPa), with a strong focus on low-level moisture (specific humidity (Q) at 850 hPa). As shown in Fig.~\ref{img444545}(b), for heavy rainfall, DPSformer's strategy shifts decisively. The importance of geopotential height diminishes across all pressure levels, suggesting less reliance on background circulation alone. Meanwhile, the focus on moisture shifts upwards, with the relative importance of specific humidity at 500 hPa increasing substantially, along with that of mid-tropospheric meridional wind (V). This reveals DPSformer has learned to operate in a manner consistent with the principles of ingredients-based forecasting. It correctly identifies that predicting extreme events requires the co-location of deep moisture and atmospheric instability, rather than relying solely on the large-scale circulation patterns represented by geopotential height~\cite{guo2024moisture, cuo2017spatial}. This finding is reinforced by a broader analysis of all variables (Supplementary Fig.~\ref{img77}), which shows that for heavy rainfall, DPSformer significantly reduces its reliance on mean sea level pressure (MSLP), instead treating the NWP precipitation forecast as an anchor to be assessed in light of the critical upper-level variables. These insights illustrate that DPSformer effectively extracts and synthesizes multiscale physical information from surface layers to the upper atmosphere—in a manner that is both physically grounded and interpretable.

\section{Discussion}\label{sec3}
Due to the inherent long-tailed distribution of rainfall data, extreme rainfall prediction presents significant challenges. From a long-tailed learning perspective, we identify the core obstacle as the model's failure to extract sufficiently discriminative and robust feature representation from rare heavy rainfall events. In this study, we introduce DPSformer to explicitly mitigate the representation deficiency caused by the long-tailed distribution. DPSformer yields a marked improvement in correcting NWP forecasts, particularly for heavy rainfall events exceeding 20~mm/6~h. For the most extreme events ($\geq$ 50~mm/6~h), DPSformer achieves a CSI of 0.067, an improvement of nearly 450\% over the original TIGGE of 0.012. This advantage becomes even more pronounced for the most widespread and high-impact rainfall events. For the top~1\% of high-coverage heavy rainfall events, the CSI improvement exceeds 50-fold, achieving an FSS greater than 0.4, which indicates a highly skillful forecast. These findings highlight the substantial potential of applying long-tailed learning to meteorological forecasting. DPSformer not only enhances prediction accuracy but also opens a promising new avenue for the development of more reliable and effective heavy rainfall forecasting systems. It serves as a valuable tool to improve disaster preparedness and mitigation efforts.

Our work is positioned at the intersection of meteorological forecasting and long-tailed learning, aiming to address a critical gap that previous methods have not adequately filled. While common strategies like reweighting loss function can emphasize rare events during training, they do not fundamentally improve the model's ability to learn the complex representation of such events \cite{hess2022deep,lin2017focal}. Furthermore, migrating general-purpose long-tailed learning strategies from computer vision presents its own challenges. We discuss the limitations of post-hoc bias-calibration methods~\cite{wang2023balancing,menon2020long} in Supplementary~Discussion~\ref{seca2}. Techniques like resampling are often inappropriate for gridded meteorological datasets, while standard data augmentation techniques struggle to generate physically plausible rainstorm patterns \cite{li2024frequency}. These highlight the need for domain-aware solutions that directly address the core challenge of representation learning from raw meteorological data.
 
Our work delivers a tangible advancement for operational weather forecasting. By correcting biases in NWP outputs, especially for high-impact events, DPSformer provides more reliable and precise guidance for meteorologists, enabling more timely and effective disaster preparedness. This hybrid approach leverages the robustness of NWP's physical constraints while harnessing AI's ability to learn complex, non-linear error patterns. This collaborative framework is poised to significantly advance the next generation of weather forecasting systems, making them more resilient and trustworthy.

Despite its promising performance, DPSformer has limitations that warrant discussion. Like many deep learning models in meteorology, it tends to generate spatially smoothed predictions, which in turn leads to over-prediction. While this bias reduces the risk of dangerous missed detections, it comes at the cost of more frequent false alarms. This trade-off must be carefully managed in operational settings. Furthermore, the reliance on TIGGE and TRMM datasets introduces a risk of regional inequity due to potential biases, such as non-uniform satellite coverage. Future work will aim to mitigate these limitations, for instance by exploring generative models to address spatial smoothing and by incorporating multi-source datasets and domain adaptation to enhance fairness and generalizability. More broadly, our results validate the framing of extreme event prediction as a long-tailed learning challenge. This conceptual framework holds significant potential to be extended to a wider range of rare, high-impact meteorological phenomena, such as tropical cyclone intensification or extreme heatwaves, opening a new frontier for forecasting and disaster mitigation.

\section{Method}\label{sec4}

\subsection{Dataset}\label{subsec41}
We use forecasts provided by ECMWF within the TIGGE archive and rainfall observations from the satellite-based TRMM 3B42 V7 product~\cite{TRMM2011}. The TIGGE dataset provides global ensemble forecasts at 6-hour intervals with a spatial resolution of 0.25° × 0.25°, initialized twice daily at 00:00 and 12:00 UTC. Our study focuses on a 64 × 64 grid region (19.5°-35.25°N, 103.5°-119.25°E), one of China's most rainfall-active and topographically complex areas, during 2007-2012, with 2007-2011 serving as training and validation data and 2012 as test data. The model inputs comprise key meteorological variables from TIGGE, including temperature, geopotential height, U/V wind components, and specific humidity at 500/700/850/925 hPa levels, alongside surface variables such as total precipitation, 2 m temperature, 10 m U/V wind components, total column water, convective available potential energy, and mean sea level pressure (detailed in Supplementary Table~\ref{tab:meteorological variables}). Ground-truth rainfall is derived from the TRMM 3B42 V7 product, which provides 3-hourly mean precipitation estimates with a spatial resolution of 0.25° × 0.25°. These were aggregated into 6-hour accumulations to align with TIGGE forecasts on the same spatial grid. For classification purposes, we categorized rainfall intensities into distinct classes using five thresholds: 0.1, 3, 10, 20, and 50~mm/6~h. This setup accentuates the long-tailed nature of rainfall data, a central challenge addressed by our proposed model.

\begin{figure}
  \centering
  \includegraphics[width=1\textwidth]{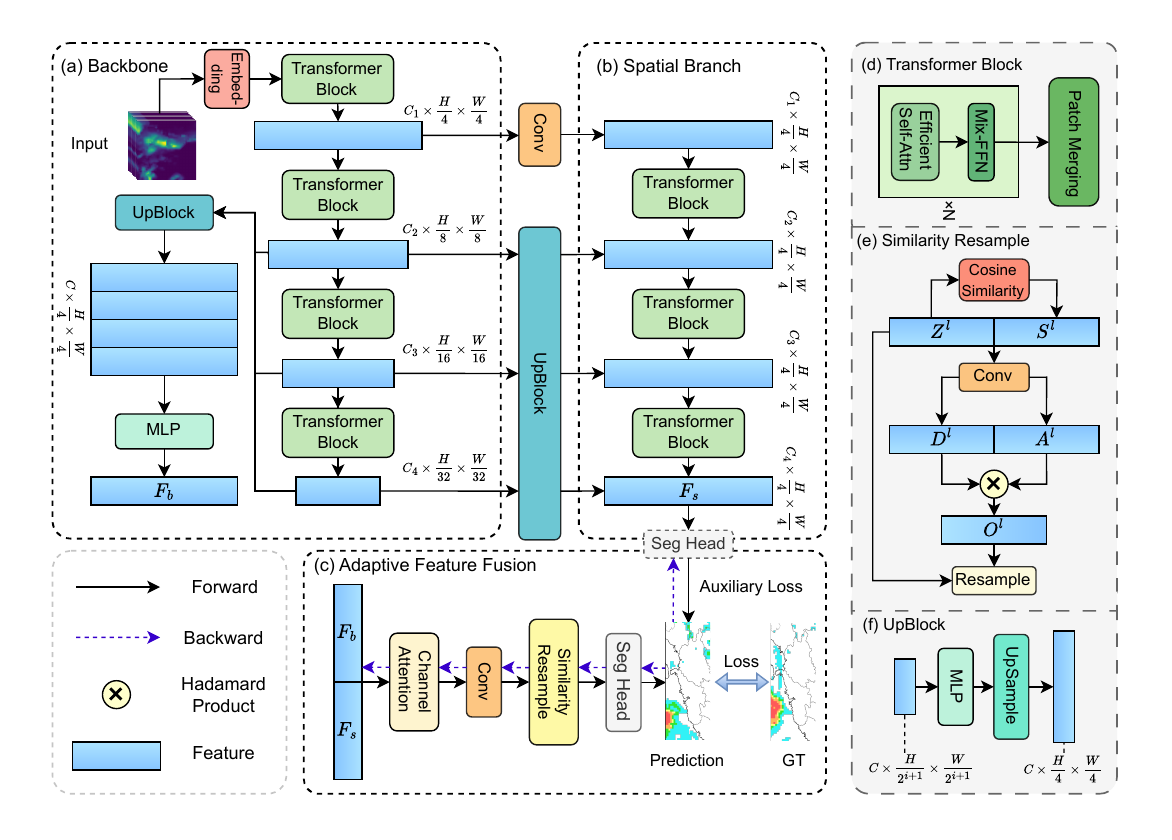} %
  \caption{\textbf{The architecture of the proposed DPSformer.} DPSformer consists of three main components: (a) Backbone for extracting general features $F_b$ from the input. (b) Spatial Branch to preserve the fine-grained spatial details, producing $F_s$ that emphasizes heavy rainfall regions. (c) Adaptive Feature Fusion module, which takes the $F_b$ and $F_s$ as inputs and intelligently integrates the information from both branches. Key sub-modules include (d) the Transformer Block, the fundamental building unit of the encoders. (e) The Similarity Resample block, which computes a similarity matrix $\mathcal{S}^l$ from the preliminarily fused features $\mathcal{Z}^l$ using cosine similarity. Offset directions $\mathcal{D}^l$ and magnitudes $\mathcal{A}^l$ are then predicted and combined into the final offsets $\mathcal{O}^l$ to guide adaptive feature resampling for fine-grained alignment. Here, $l$ denotes the feature level. (f) the UpBlock used for feature decoding.} 
  \label{img6} 
\end{figure}  

\subsection{Framework}\label{subsec42}
To address the inadequate feature representation of heavy rainfall events in models, a consequence of the long-tailed distribution of rainfall, we propose the DPSformer architecture as shown in Fig.~\ref{img6}. Standard vision transformers, such as SegFormer, acquire global context via multi-level downsampling. However, for rainfall forecasting, this process leads to an irreversible loss of fine-grained spatial details~\cite{li2024ewt}, which are critical for capturing the highly localized structures of intense convective storms~\cite{han2024fengwu}. Accordingly, DPSformer is designed with a dual-branch architecture: a multi-scale backbone for global context and a high-resolution spatial branch for local fidelity. The backbone branch, which is built upon the SegFormer-b0 architecture~\cite{xie2021segformer}, excels at modeling large-scale atmospheric patterns associated with common, lighter rainfall. In parallel, the spatial branch employs the same Transformer Block modules as the backbone but maintains a fixed, higher-resolution pathway, thereby preserving the precise location and intensity information of small-scale features. This branch functions as an expert pathway for the tail class of the data distribution, guided by a specialized loss function (see Loss function) to compensate for the information dilution that heavy rainfall events suffer in the backbone. Features from both branches are adaptively merged by a fusion module (see Adaptive feature fusion), preserving robust global discrimination while substantially enhancing the spatial representation of extreme rainfall, as validated by our feature quality analysis (see Supplementary Table~\ref{tab:feature_metrics_appendix}).

This architecture reduces the imbalance in feature learning between common and rare events and mitigates the representation deficiency arising from long-tailed rainfall distributions. It embodies a key principle of long-tailed learning theory: enhancing the representation of minority classes.

\subsection{Adaptive feature fusion}\label{subsec43}
To mitigate the challenge of representation deficiency for rare, high-intensity rainfall classes, we propose a novel adaptive feature fusion module. This module is designed to efficiently integrate information from the backbone and spatial branches. It leverages information from the spatial branch, which is highly correlated with heavy rainfall, to address the backbone's under-representation of tail-end categories. The process begins by concatenating features from both branches to leverage multi-scale information, followed by a channel attention~\cite{hu2018squeeze} that re-weights channels to prioritize task-relevant information. This initial fusion is performed using convolution operations.

However, direct fusion often leads to class inconsistency and boundary displacement, particularly for rainfall fields, which exhibit fuzzy, drifting boundaries that make static fusion inherently misaligned. To overcome this difficulty, we introduce an offset generator~\cite{chen2024frequency} that performs adaptive feature resampling for fine-grained alignment. This mechanism is motivated by the need to flexibly align features across scales, ensuring that strong, stable feature regions can inform and correct weaker, inconsistent ones, especially at critical class boundaries. The generator first computes a similarity matrix $\mathcal{S}^l$ by measuring the cosine similarity between each pixel in the initially fused feature map $\mathcal{Z}^l$ and its eight neighboring pixels. Here, $l$ denotes the feature level. Let $H$ and $W$ be the spatial dimensions of the feature map. Given $\mathcal{Z}^l$ and $\mathcal{S}^l$, the offsets are predicted as follows:
\begin{align}
  \mathcal{D}^l &= Conv_{3 \times 3}(Concat(\mathcal{Z}^l, \mathcal{S}^l)), \\
  \mathcal{A}^l &= Sigmoid(Conv_{3 \times 3}(Concat(\mathcal{Z}^l, \mathcal{S}^l))), \\
  \mathcal{O}^l &= \mathcal{D}^l \cdot \mathcal{A}^l,
\end{align}
where $\mathcal{O}^l \in \mathbb{R}^{2G \times H \times W}$ denotes the final predicted offsets, $\mathcal{D}^l$ represents offset directions, and $\mathcal{A}^l$ corresponds to offset magnitudes. $\operatorname{Concat}$ denotes the channel-wise concatenation operation, and $\operatorname{Sigmoid}$ is the sigmoid activation function. The feature map is divided into $G$ distinct groups, each assigned unique spatial offsets. This grouping strategy enables differentiated resampling, allowing features with high intra-category coherence to replace inconsistent ones, thereby refining class boundaries and enhancing the robustness of the final prediction.

\subsection{Loss function}\label{subsec44}
Heavy rainfall prediction requires the concurrent optimization of spatial localization and intensity estimation. In this context, the CE loss is commonly employed to enhance intensity classification by penalizing deviations in predicted class probabilities, whereas the Dice loss directly optimizes spatial overlap between predicted and observed rainfall regions. In our formulation, the Dice loss is computed exclusively for heavy rainfall categories by merging the first four rainfall classes into a single background class and retaining only the last two classes as positive targets. These complementary objectives are assigned to both the spatial branch and the main branch, with their contributions combined into the overall loss:
\begin{equation}
\mathcal{L}_{\mathrm{overall}} = (1 - \alpha) \cdot \mathcal{L}_{\mathrm{main}} + \alpha \cdot \mathcal{L}_{\mathrm{spatial}},
\end{equation}
where $\alpha \in [0,1]$ controls the relative contribution of the two branches, $\mathcal{L}_{\mathrm{main}}$ denotes the loss computed from the model's final output, and $\mathcal{L}_{\mathrm{spatial}}$ represents an auxiliary loss designed to ensure that the spatial branch focuses on capturing spatial information relevant to heavy rainfall.

The spatial branch is designed to preserve high-resolution feature maps while capturing spatial patterns specifically associated with heavy rainfall. Its loss combines a Dice loss with a WCE loss enhanced by the Balancing Logit Variation (BLV) method~\cite{wang2023balancing}, with a higher weight assigned to the Dice component ($\gamma > 0.5$):
\begin{equation}
\mathcal{L}_{\mathrm{spatial}} = \gamma \cdot \mathcal{L}_{\mathrm{Dice}} + (1 - \gamma) \cdot \mathcal{L}_{\mathrm{WCE}}(\hat{z}),
\end{equation}
where $\mathcal{L}_{\mathrm{WCE}}(\hat{z})$ denotes the WCE loss computed on BLV-perturbed logits. The weights for the WCE are calculated using the square root of the inverse of each class's frequency. BLV mitigates the impact of class imbalance by adding class-frequency-weighted Gaussian noise to the logits $z$:
\begin{equation}
\hat{z}_{i}^{k} = z_{i}^{k} + \frac{c_{k}}{\max_{i=0}^{C-1} c_{j}} \, \mathcal{N}\left(0, \sigma^{2}\right), 
\quad 
c_{k} = \log \left( \frac{\sum_{j=0}^{C-1} q_{j}}{q_{k}} \right),
\end{equation}
where $i$ denotes the $i$-th pixel in the logits $z$. $q_{k}$ is the sample count for class $k$ and $\sigma$ is the standard deviation of the Gaussian noise. This perturbation magnitude is inversely related to class frequency, encouraging balanced representation learning in the feature space. Optimizing the spatial branch in this manner enables the model to learn discriminative and spatially calibrated features in the earlier stage.

Building upon these enhanced features, the main branch loss operates on the final output and combines a CE loss and a Dice loss:
\begin{equation}
\mathcal{L}_{\mathrm{main}} = (1 - \gamma) \cdot \mathcal{L}_{\mathrm{CE}}(\tilde{z}) + \gamma \cdot \mathcal{L}_{\mathrm{Dice}}.
\end{equation}
Before computing the CE, a logit adjustment (LA) strategy~\cite{menon2020long} is applied to further address class imbalance:
\begin{equation}
\tilde{z}_{i}^{k} = z_{i}^{k} + \tau \cdot \log \pi_k,
\end{equation}
where $\pi_k$ denotes the empirical prior probability of class $k$, and $\tau$ is a scaling hyperparameter. As shown in Supplementary Table~\ref{tab:feature_metrics_appendix} for the migrating long-tailed learning experiment, applying LA in isolation produced limited benefit; however, when combined with the spatial-branch optimization, LA leverages the resulting balanced and spatially precise features to more effectively shift the decision boundary towards minority classes without degrading head-class performance.

Overall, the training scheme first optimizes the spatial branch with Dice and BLV-enhanced WCE to establish spatially precise and balanced features, then applies CE and Dice in the main branch, and finally incorporates LA to refine the decision boundaries. This sequence ensures that each component operates on appropriately calibrated features, thereby improving both spatial localization and intensity estimation in heavy rainfall prediction.

\backmatter

\bmhead{Data availability}
The datasets used in this study are publicly available. The TIGGE data were obtained from the ECMWF data portal (\url{https://apps.ecmwf.int/datasets/data/tigge/}). The TRMM 3B42 dataset was sourced from the NASA Goddard Earth Sciences Data and Information Services Center (GES DISC) repository (\url{https://disc.gsfc.nasa.gov/datasets/TRMM_3B42_7/summary}).

\bmhead{Author contribution}
Z.W. and Y.L. supervised the project. Z.W., Y.L., and T.S. proposed the initial concept. Z.H. designed the methodology and implemented the model. Z.H. and T.S. wrote the original manuscript. Z.H., T.S., Z.W., Y.L., and Y.Y. contributed to the subsequent revisions. W.Z. and H.W. were responsible for project and resource management.

\bmhead{Competing interests}
The authors declare no competing interests.


\clearpage  

\setcounter{equation}{0}
\setcounter{figure}{0}
\setcounter{table}{0}
\setcounter{section}{0}
\renewcommand{\theequation}{S\arabic{equation}}
\renewcommand{\thefigure}{S\arabic{figure}}
\renewcommand{\thetable}{S\arabic{table}}
\renewcommand{\thesection}{S\arabic{section}}
\renewcommand{\bibnumfmt}[1]{[S#1]} 

\onecolumn 
\begin{center}
\textbf{\Large Supplementary Information for \\ DPSformer: A long-tail-aware model for improving heavy rainfall prediction}
\end{center}

\section{Migrating long-tailed learning}\label{seca2}
In this section, we directly transfer representative long-tailed learning methods to the rainfall forecasting task. The complete results are presented in Table~\ref{tab:detailed_results}. We focus on evaluating two representative long-tailed strategies designed to calibrate the output bias toward rare samples: LA and BLV. These methods, proven effective on natural images, aim to mitigate the classifier's output bias against rare samples. However, as shown in the Table~\ref{tab:detailed_results} for 20~mm/6~h and 50~mm/6~h thresholds, their CSI scores do not significantly improve compared to the standard SegFormer backbone. At 50~mm/6~h thresholds, SegFormer achieves a CSI of 0.050, while LA drops to 0.045. At 20~mm/6~h thresholds, LA and BLV (0.151 and 0.155) fail to match SegFormer's 0.156. 

A key reason for this systematic failure is that most long-tailed learning algorithms (e.g. LA) assume an imbalanced training distribution but a balanced test distribution. This assumption motivates the design of cost-sensitive training objectives or output calibration techniques. However, in the task of rainfall forecasting, both training and testing distributions exhibit a severe long-tailed nature. Consequently, when applied, these methods often overcompensate for rare samples, leading to performance degradation or unstable predictions. As shown in Table~\ref{tab:detailed_results}, all transferred long-tailed learning methods result in a notable decrease in POD; for instance, the POD of LA drops from SegFormer's 0.126 to 0.102.

Furthermore, bias-calibration methods are based on a potentially invalid assumption: that the long-tailed data distribution does not significantly affect the model's ability to extract features for rare events. For complex meteorological data, this assumption may not hold true. Performing bias calibration with insufficiently learned features for heavy rainfall can be counterproductive. To quantitatively test this hypothesis, we conducted a feature quality analysis (Table~\ref{tab:feature_metrics_appendix}) to assess the separability of the learned representations. The results confirm that these methods struggle at the feature level. LA and BLV achieve very low CH scores (2.4–2.5) and Fisher scores ($<$ 0.1) for heavy rainfall events, showing almost no improvement over the original SegFormer at the feature representation level.

In summary, a significant mismatch exists between the assumptions of classical long-tailed learning methods and the characteristics of rainfall forecasting, rendering output calibration alone insufficient to predict extreme rainfall events effectively. Given the long-tailed distribution of precipitation, enabling the model to learn effective features for rare heavy rainfall events remains a key bottleneck in improving forecasting performance. Feature-level evaluation in Table~\ref{tab:feature_metrics_appendix} confirms that our DPSformer achieves the highest CH (19.7) and Fisher (0.363) scores, far surpassing all bias-calibration baselines. These strong feature representations directly translate into superior predictive performance, with DPSformer delivering the highest CSI across two heavy rainfall thresholds and achieving substantial gains over the SegFormer backbone.

\begin{table*}[t]
  \centering
  \begin{threeparttable}
  \caption{\textbf{Performance of baseline and long‑tail methods for two rainfall thresholds.} The last four methods (Resample, LA, FocalLoss, BLV) are long-tail methods, all trained with SegFormer as the backbone. The SegFormer model capacity used is the SegFormer-b0 variant.
 The ideal value of Bias is 1. FSS is calculated using a neighborhood size of 5 grid points.}

  \label{tab:detailed_results}
  \small            
  \setlength{\tabcolsep}{3.3pt}  
  \renewcommand{\arraystretch}{1.15}

  \begin{tabular}{lccccccccc}
  \toprule
  \multicolumn{10}{c}{$\geq 20$ mm / 6 h} \\
  \midrule
  Method & CSI\,\raisebox{+0.25ex}{$\uparrow$} & ETS\,\raisebox{+0.25ex}{$\uparrow$} & POD\,\raisebox{+0.25ex}{$\uparrow$} & Bias & MAR\,\raisebox{+0.25ex}{$\downarrow$} & FAR\,\raisebox{+0.25ex}{$\downarrow$} & F1\,\raisebox{+0.25ex}{$\uparrow$} & $\Delta$SEDI\,\raisebox{+0.25ex}{$\uparrow$} & FSS\,\raisebox{+0.25ex}{$\uparrow$} \\ 
  \midrule
  TIGGE      & 0.070 & 0.067 & 0.084 & 0.297 & 0.916 & \textbf{0.716} & 0.106 & --    & 0.108 \\
  QM         & 0.119 & 0.113 & 0.220 & \textbf{1.063} & 0.780 & 0.793 & 0.168 & 0.065 & 0.328 \\
  U\text{-}Net    & 0.113 & 0.104 & \underline{0.522} & 4.253 & \underline{0.478} & 0.870 & 0.147 & 0.269 & 0.263 \\
  SegFormer  & 0.156 & 0.149 & 0.343 & 1.543 & 0.657 & 0.776 & 0.104 & 0.212 & 0.393 \\
  WMSE\text{-}MS\text{-}SSIM & 0.127 & 0.118 & \textbf{0.526} & 3.778 & \textbf{0.474} & 0.855 & 0.109 & \textbf{0.314} & 0.269 \\
  OBDice     & \underline{0.165} & \underline{0.159} & 0.320 & \underline{1.254} & 0.680 & \underline{0.741} & \underline{0.179} & 0.226 & \underline{0.431} \\
  Resample   & 0.142 & 0.135 & 0.344 & 1.770 & 0.656 & 0.803 & 0.136 & 0.180 & 0.365 \\
  LA         & 0.151 & 0.145 & 0.336 & 1.553 & 0.664 & 0.783 & 0.134 & 0.196 & 0.388 \\
  FocalLoss  & 0.157 & 0.149 & 0.377 & 1.785 & 0.623 & 0.788 & 0.092 & 0.236 & 0.385 \\
  BLV        & 0.165 & 0.148 & 0.332 & 1.479 & 0.668 & 0.773 & 0.115 & 0.203 & 0.395 \\
  DPSformer & \textbf{0.174} & \textbf{0.167} & 0.382 & 1.579 & 0.618 & 0.757 & \textbf{0.181} & \underline{0.280} & \textbf{0.433} \\
  \midrule
  \multicolumn{10}{c}{$\geq 50$ mm / 6 h} \\
  \midrule
  Method & CSI\,\raisebox{+0.25ex}{$\uparrow$} & ETS\,\raisebox{+0.25ex}{$\uparrow$} & POD\,\raisebox{+0.25ex}{$\uparrow$} & Bias & MAR\,\raisebox{+0.25ex}{$\downarrow$} & FAR\,\raisebox{+0.25ex}{$\downarrow$} & F1\,\raisebox{+0.25ex}{$\uparrow$} & $\Delta$SEDI\,\raisebox{+0.25ex}{$\uparrow$} & FSS\,\raisebox{+0.25ex}{$\uparrow$} \\ 
  \midrule
  TIGGE      & 0.012 & 0.012 & 0.013 & 0.111 & 0.987 & \textbf{0.881} & 0.024 & --    & 0.008 \\
  QM         & 0.031 & 0.030 & 0.050 & 0.672 & 0.950 & 0.926 & 0.060 & 0.003 & 0.112 \\
  U\text{-}Net    & 0.029 & 0.028 & \textbf{0.327} & 11.07 & 0.672 & 0.965 & 0.057 & \textbf{0.167} & 0.102 \\
  SegFormer  & 0.050 & 0.049 & 0.126 & 1.617 & 0.873 & 0.920 & 0.096 & 0.049 & 0.160 \\
  WMSE\text{-}MS\text{-}SSIM & 0.048 & 0.047 & 0.170 & 2.781 & \underline{0.830} & 0.932 & 0.092 & 0.066 & 0.126 \\
  OBDice     & \underline{0.056} & \underline{0.055} & 0.108 & \textbf{1.045} & 0.892 & \underline{0.895} & \underline{0.106} & 0.053 & \textbf{0.199} \\
  Resample   & 0.044 & 0.043 & 0.111 & 1.680 & 0.888 & 0.930 & 0.084 & 0.033 & 0.145 \\
  LA         & 0.045 & 0.044 & 0.102 & 1.353 & 0.898 & 0.923 & 0.086 & 0.032 & 0.151 \\
  FocalLoss  & 0.051 & 0.051 & 0.113 & \underline{1.309} & 0.887 & 0.910 & 0.098 & 0.065 & 0.171 \\
  BLV        & 0.051 & 0.050 & 0.123 & 1.479 & 0.877 & 0.918 & 0.098 & 0.048 & 0.162 \\
  DPSformer & \textbf{0.067} & \textbf{0.066} & \underline{0.220} & 2.483 & \textbf{0.780} & 0.911 & \textbf{0.126} & \underline{0.117} & \underline{0.197} \\
  \bottomrule
  \end{tabular}

  \begin{tablenotes}[flushleft]
  \footnotesize
  \item \raisebox{+0.25ex}{$\uparrow$} Indicates that higher values are better for the corresponding metric.
  \item \raisebox{+0.25ex}{$\downarrow$} Indicates that lower values are better for the corresponding metric.
  \item The best result in each column is shown in \textbf{bold}, and the second-best is \underline{underlined}.
  \end{tablenotes}

  \end{threeparttable}
\end{table*}

\begin{table}[ht] 
\small   
\centering 
\caption{\textbf{Feature Quality Result.} 
Higher values indicate better separability. For KNNAcc, the number of neighbors (K) is set to 1. The SegFormer model capacity used is the SegFormer-b0 variant.We report results for four representative long-tailed learning methods (Resample, LA, FocalLoss, BLV) and the strongest meteorology-specific method (OBDice), showing that these approaches cannot fully overcome the feature deficiency issue. The features from each component of our model consistently exhibit superior quality.} 
\small            
\setlength{\tabcolsep}{4pt}  
\label{tab:feature_metrics_appendix} 
\begin{tabular}{ 
  @{}l 
  S[table-format=1.3]  
  S[table-format=1.3]  
  S[table-format=1.3]  
  S[table-format=1.3]  
  S[table-format=2.1]  
  S[table-format=1.3]  
} 
\toprule
Method &
{IntraSim\,\raisebox{+0.25ex}{$\uparrow$}} &
{SimMargin\,\raisebox{+0.25ex}{$\uparrow$}} &
{KNNAcc\,\raisebox{+0.25ex}{$\uparrow$}} &
{Fisher\,\raisebox{+0.25ex}{$\uparrow$}} &
{$\mathrm{CH}^{\ast}$\,\raisebox{+0.25ex}{$\uparrow$}} &
{$F1_{\mathrm{SVM}}^{\ast}$\,\raisebox{+0.25ex}{$\uparrow$}} \\
\midrule
SegFormer & 0.374 & 0.004 & 0.382 & 0.008 & 2.4 & 0.312 \\
OBDice          & 0.400 & 0.006 & 0.393 & 0.009 & 2.5 & 0.387 \\
Resample        & 0.384 & 0.008 & 0.383 & 0.009 & 2.3 & 0.303 \\
LA              & \underline{0.626} & 0.001 & 0.384 & 0.005 & 2.5 & 0.293 \\
FocalLoss       & 0.300 & 0.009 & 0.391 & 0.010 & 2.6 & 0.313 \\
BLV             & 0.450 & 0.004 & 0.397 & 0.009 & 2.4 & 0.304 \\
\midrule
DPSformer (Backbone) & 0.489 & 0.037 & 0.393 & 0.210 & 4.8 & \textbf{0.463} \\
DPSformer (Spatial)  & 0.544 & \underline{0.102} & \underline{0.398} & \underline{0.227} & \underline{18.7} & 0.406 \\
DPSformer   & \textbf{0.647} & \textbf{0.118} & \textbf{0.402} & \textbf{0.363} & \textbf{19.7} & \underline{0.429} \\
\bottomrule
\end{tabular} 
\vspace{1.5mm} 
\parbox{11.5cm}{ 
\footnotesize
\raisebox{+0.25ex}{$\uparrow$} Indicates that higher values are better for the corresponding metric. \\
\raisebox{+0.25ex}{$\ast$}~Metrics are computed on heavy rainfall samples only. \\
The best result in each column is shown in \textbf{bold}, and the second‑best is \underline{underlined}.
}
\end{table}

\section{Ablations}\label{seca3}
To evaluate the contributions of each design component in DPSformer, we conducted an ablation study under the 50~mm/6~h heavy rainfall threshold, with the results presented in Table~\ref{tab:ablation_50mm}. Starting with the SegFormer backbone, we incrementally incorporated high-resolution features (HRF), a dual-path architecture, and a custom Dual Loss function tailored for heavy rainfall. First, replacing the progressively downsampled features in the original SegFormer with constant high-resolution representations yielded an improvement in overall performance. The CSI increased from 0.050 to 0.055, and the F1-score rose from 0.096 to 0.104, while the Bias decreased from 1.617 to 1.452. This indicates that preserving finer spatial detail helps the model achieve more balanced predictions. Next, building upon this foundation, we introduced the dual-path architecture. The synergy between the main backbone and a new spatial branch significantly enhanced the model's ability to capture heavy rainfall events, boosting the POD from 0.127 to 0.170. Although this led to a higher overforecasting tendency (Bias increased from 1.452 to 1.993), key composite metrics such as CSI (0.060) and F1-score (0.113) continued to improve, confirming the effectiveness of the dual-path design in identifying more intense rainfall events. Finally, the fully integrated DPSformer, equipped with all three components, delivered superior performance across the board. It achieved peak scores in CSI (0.067), POD (0.220), and F1-score (0.126). These results conclusively demonstrate that each proposed component positively contributes to the final performance, and their combination is most effective for forecasting rare and intense heavy rainfall events.

\begin{table}[!t]
\centering
\caption{\textbf{Ablation study on HRF, Dual Path, and Dual Loss components for the 50~mm/6~h heavy rainfall threshold.} The ideal value of Bias is 1.}
\label{tab:ablation_50mm}
\setlength{\tabcolsep}{4pt}  
\begin{tabularx}{\linewidth}{
    >{\centering\arraybackslash}p{0.5cm} 
    >{\centering\arraybackslash}p{1.2cm} 
    >{\centering\arraybackslash}p{1.2cm}
    *{7}{>{\centering\arraybackslash}X}
}
\toprule
\multirow{1}{*}{HRF} & \multirow{1}{*}{Dual Path} & \multirow{1}{*}{Dual Loss} 
& CSI\,\raisebox{+0.25ex}{$\uparrow$} 
& POD\,\raisebox{+0.25ex}{$\uparrow$} 
& FAR\,\raisebox{+0.25ex}{$\downarrow$} 
& Bias 
& F1\,\raisebox{+0.25ex}{$\uparrow$} 
& $\Delta$SEDI\,\raisebox{+0.25ex}{$\uparrow$} 
& FSS\,\raisebox{+0.25ex}{$\uparrow$} \\
\midrule
\ding{55} & \ding{55} & \ding{55} & 0.050 & 0.126 & 0.920 & \underline{1.617} & 0.096 & 0.049 & 0.160 \\
\ding{51} & \ding{55} & \ding{55} & 0.055 & 0.127 & \underline{0.912} & \textbf{1.452} & 0.104 & 0.056 & 0.152 \\
\ding{51} & \ding{51} & \ding{55} & \underline{0.060} & \underline{0.170} & 0.914 & 1.993 & \underline{0.113} & \underline{0.081} & \underline{0.164} \\
\ding{51} & \ding{51} & \ding{51} & \textbf{0.067} & \textbf{0.220} & \textbf{0.911} & 2.483 & \textbf{0.126} & \textbf{0.117} & \textbf{0.197} \\
\bottomrule
\end{tabularx}
\vspace{1.5mm} 
\parbox{11.5cm}{ 
\footnotesize
\raisebox{+0.25ex}{$\uparrow$} Indicates that higher values are better for the corresponding metric. \\
 \raisebox{+0.25ex}{$\downarrow$} Indicates that lower values are better for the corresponding metric. \\
 \ding{55}/\ding{51}~Denote inclusion/exclusion of the corresponding component. \\
The best result in each column is shown in \textbf{bold}, and the second‑best is \underline{underlined}.
}
\end{table}

To verify that the performance enhancement stems from architectural innovation rather than merely an increased parameter count, we compared DPSformer (16.93M parameters) with SegFormer-b1 (13.76M), a model with a similar scale, and SegFormer-b2 (27.43M), a much larger model. As shown in Table \ref{tab:Parameter size}, despite having significantly fewer parameters than SegFormer-b2, DPSformer demonstrates markedly superior performance across all forecast metrics (CSI: 0.067 vs. 0.046; $\Delta$SEDI: 0.117 vs. 0.046). This result provides strong evidence that the unique dual-branch architecture and loss function of DPSformer are the key drivers of its performance, not simply an increase in model capacity. Further insight comes from the quantitative evaluation of heavy rainfall events' feature quality using CH and $F1_{\mathrm{SVM}}$. DPSformer achieves a CH score of 19.7, nearly an order of magnitude higher than the baselines (2.0–2.8). This striking gap shows that our dual‐branch design enables the learning of more discriminative and robust representations for rare heavy‐rainfall events. In contrast, even the largest backbone, SegFormer‐b2, fails to improve feature quality (CH =2.8) meaningfully, explaining its limited forecasting performance. Finally, our experiments reveal that a high-quality feature foundation is a prerequisite for the effectiveness of classification bias correction methods, such as LA. When the LA method was applied to the SegFormer models, which exhibit poor feature quality, it yielded only marginal gains and even led to performance degradation on SegFormer-b2 (CSI decreased from 0.046 to 0.044). This demonstrates that when a model's underlying feature extraction capability is insufficient, classifier-level adjustments are ineffective. Only when a model, such as DPSformer, first acquires a high-quality feature basis through architectural innovation can subsequent correction methods effectively optimize predictions and realize their full potential.

\begin{table}[htbp]
    \centering
    \caption{\textbf{The impact of parameter number on model performance.}
    The ideal value of Bias is 1.}
    \label{tab:Parameter size} 
    \begin{tabular}{lccccccc} 
        \toprule 
        
        \multirow{2}{*}{Method} & \multicolumn{4}{c}{$\ge 50$~mm/6~h} & \multicolumn{2}{c}{Feature Quality} & \multirow{2}{*}{Parameters} \\
        
        \cmidrule(lr){2-5} \cmidrule(lr){6-7}
        & CSI\,\raisebox{+0.25ex}{$\uparrow$} & POD\,\raisebox{+0.25ex}{$\uparrow$} & Bias & $\Delta$SEDI\,\raisebox{+0.25ex}{$\uparrow$} & {$\mathrm{CH}^{\ast}$}\,\raisebox{+0.25ex}{$\uparrow$} & {$F1_{\mathrm{SVM}}^{\ast}$}\,\raisebox{+0.25ex}{$\uparrow$} & \\
        
        \midrule 
        
        SegFormer b0 (WCE)& 0.050 & 0.126 & \textbf{1.617} & 0.049 & 2.5 & 0.393 & 3.75M \\
        SegFormer b1 (WCE) & 0.050 & \underline{0.182} & 2.794 & \underline{0.072} & 2.2 & \underline{0.435} & 13.76M \\
        SegFormer b2 (WCE) & 0.046 & 0.131 & \underline{1.914} & 0.046 & \underline{2.8} & 0.434 & 27.43M \\
        SegFormer b1 (LA) & \underline{0.051} & 0.160 & 2.335 & 0.062 & 2.0 & \textbf{0.436} & 13.76M \\
        SegFormer b2 (LA) & 0.044 & 0.133 & 2.139 & 0.041 & 2.6 & 0.386 & 27.43M \\
        DPSformer & \textbf{0.067} & \textbf{0.220} & 2.483 & \textbf{0.117} & \textbf{19.7} & 0.429 & 16.93M \\
        
        \bottomrule 
    \end{tabular}
    \vspace{1.5mm} 
\parbox{11.5cm}{ 
\footnotesize
\raisebox{+0.25ex}{$\uparrow$} Indicates that higher values are better for the corresponding metric. \\
\raisebox{+0.25ex}{$\ast$}~Metrics are computed on heavy rainfall samples only. \\
The best result in each column is shown in \textbf{bold}, and the second‑best is \underline{underlined}.
}
\end{table}

\section{Interpretation}\label{seca4}
To interpret the contribution of each meteorological input variable to the predictions, we employ Integrated Gradients (IG). For an input tensor $\mathbf{x}$ and a baseline $\mathbf{x}'$, the attribution for feature $i$ is:
\begin{equation}
\label{eq:ig}
\mathrm{IG}_i(\mathbf{x}) \;=\; (x_i - x'_i)\;\times\; \int_{0}^{1} 
\frac{\partial f(\mathbf{x}_\alpha)}{\partial x_i}\, d\alpha,
\quad\text{with}\quad 
\mathbf{x}_\alpha \,=\, \mathbf{x}' + \alpha\,(\mathbf{x}-\mathbf{x}'),
\end{equation}
where $f(\cdot)$ is a scalar-valued function of the model output.

Because DPSformer produces multi-class, pixel-wise logits, we reduce them to a scalar by spatially averaging logits either (i) over all pixels (global importance) or (ii) over pixels belonging to the heavy rainfall classes (Level 4 and 5). IG computed with this scalar $f(\cdot)$ yields a channel-wise importance score per sample. We use a zero-valued input as the default baseline. Per-sample attributions are then averaged across the evaluation set to obtain a single relative importance score for each of the 27 meteorological variables. 

\begin{figure}
  \centering
  \includegraphics[width=1.0\textwidth]{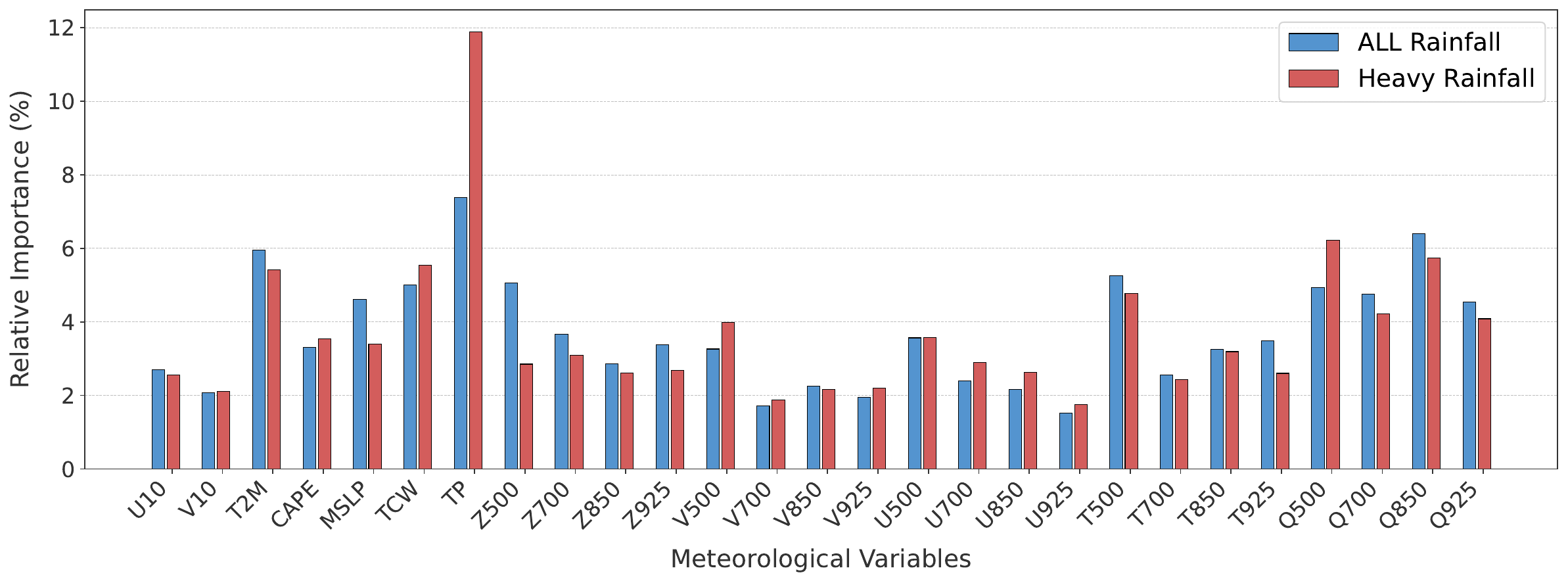} 
  \caption{\textbf{IG attribution importance for each meteorological variable.} U/V10: 10-metre wind components; T2M: 2-metre temperature; CAPE: convective available potential energy; MSLP: mean sea level pressure; TCW: total column water; TP: original NWP accumulated precipitation; ZXXX: geopotential height at different pressure levels; U/VXXX: U/V wind components at different pressure levels; TXXX: temperature at different pressure levels(500/700/850/925 hPa); QXXX: specific humidity at different pressure levels. The model relies more on upper-level and column-integrated moisture variables under heavy rainfall, whereas for all pixels, it places greater emphasis on surface pressure, low-level circulation, and humidity.} 
  \label{img77} 
\end{figure} 

As seen in Fig. \ref{img77}, the IG attribution results demonstrate that DPSformer's ranking of key variables is highly consistent with established meteorological theory. For general rainfall (blue bars), the model emphasizes the NWP original forecast (TP) as the dominant factor, supplemented by large-scale circulation patterns via geopotential height (Z500) and low-level moisture (Q850). Low-level winds (U10, V10) and surface variables like T2M and MSLP also contribute moderately, reflecting a balanced reliance on near-surface dynamics and moisture availability. In contrast, for heavy rainfall (red bars), TP's importance surges dramatically, indicating its role as a critical anchor for extreme events. The model shifts toward an enhanced focus on moisture, with upper-level specific humidity (Q500) gaining prominence, alongside enhanced weighting of meridional winds (V500, V700) that control deep moisture transport. Meanwhile, the relative role of geopotential height (Z500) and MSLP diminishes, suggesting that during heavy rainfall, large-scale background circulation becomes secondary to deep-layer moisture convergence and vertical lifting. This adaptive structure aligns closely with classic heavy rainfall theories. These theories emphasize the interplay of abundant moisture, dynamic lifting, and conditional instability. The model's learned strategy is consistent with quantitative findings from previous studies and recent circulation-based attribution research. This alignment confirms the model's physical consistency, interpretability, and scientific reliability.

\section{Evaluation metrics}\label{seca5}

\subsubsection*{Rainfall performance metrics}\label{subseca51}
To comprehensively evaluate the model's rainfall forecasting performance, we adopted the following verification metrics, which are derived from a confusion matrix (Table \ref{tab:confusion_matrix}):

\begin{table}[h!]
\centering
\caption{\textbf{Confusion matrix for forecast verification.} The matrix categorizes forecast outcomes based on observed events. This matrix serves as the basis for the categorical verification metrics employed in this study.}
\label{tab:confusion_matrix}
\renewcommand{\arraystretch}{1.2} 
\begin{tabular}{@{}lll@{}}
\toprule
 & \textbf{Forecast Event} & \textbf{Forecast No Event} \\ 
\midrule
\textbf{Observed Event} & TP (Hits) & FN (Misses) \\
\textbf{Observed No Event} & FP (False Alarms) & TN (Correct Rejections) \\ 
\bottomrule
\end{tabular}
\end{table}

\textbf{Critical Success Index (CSI)} measures how well forecast events correspond to observed events. It is widely used for rare event verification because it ignores true negatives:
\begin{equation}
  \text{CSI} = \frac{TP}{TP + FN + FP},
  \label{eq:csi}
\end{equation}
where TP, FN, and FP are true positives, false negatives, and false positives. The score penalizes both misses and false alarms, providing a balanced measure of event prediction accuracy. Its range is $[0, 1]$, with 1 being perfect.

\textbf{Equitable Threat Score (ETS)} adjusts the CSI for hits that would occur by random chance. It evaluates the forecast skill relative to a random baseline:
\begin{equation}
  \text{ETS} = \frac{TP - TP_{random}}{TP + FN + FP - TP_{random}},
  \label{eq:ets}
\end{equation}
where $TP_{random}$ quantifies hits from random chance. By removing these, ETS provides a fair skill measure. A score of 0 indicates no skill over random, while 1 is a perfect forecast. Its range is $[-1/3, 1]$.

\textbf{Probability of Detection (POD)} measures the fraction of observed events that were correctly forecast:
\begin{equation}
  \text{POD} = \frac{TP}{TP + FN}.
  \label{eq:pod}
\end{equation}
Pod is a direct measure of the model's ability to detect events, but it ignores false alarms. Therefore, it can be artificially inflated and should be assessed alongside the FAR. The range is $[0, 1]$, with 1 being perfect.

\textbf{Bias} score is the ratio of the frequency of forecast events to the frequency of observed events. It indicates systematic overforecasting or underforecasting:
\begin{equation}
  \text{Bias} = \frac{TP + FP}{TP + FN}.
  \label{eq:bias}
\end{equation}
A score $>1$ indicates overforecasting, $<1$ indicates underforecasting, and $=1$ indicates no frequency bias.

\textbf{Miss Alarm Ratio (MAR)} is the fraction of observed events that the forecast failed to predict. It is the complement of POD:
\begin{equation}
  \text{MAR} = \frac{FN}{TP + FN}.
  \label{eq:mar}
\end{equation}
MAR provides a direct measure of prediction failures for actual events, where $\text{MAR} = 1 - \text{POD}$. A lower value is better, with 0 being a perfect score.

\textbf{False Alarm Ratio (FAR)} is the fraction of predicted events that did not actually occur; it measures the forecast's propensity to incorrectly indicate the presence of an event:
\begin{equation}
  \text{FAR} = \frac{FP}{TP + FP}.
  \label{eq:far}
\end{equation}
FAR quantifies forecast inefficiency due to false alarms and is related to Precision ($1 - \text{FAR}$). A lower value is better, with 0 indicating that every event prediction was correct.

\textbf{F1} score is the harmonic mean of Precision and Recall (POD). It provides a single, balanced summary of model performance, especially for imbalanced datasets:
\begin{equation}
  \text{F1} = \frac{2TP}{2TP + FP + FN}.
  \label{eq:f1}
\end{equation}
As a harmonic mean, F1 score is high only when both precision (low FAR) and recall (high POD) are high. This makes it a more stringent metric than the arithmetic mean. Its range is $[0, 1]$, with 1 being best.

\textbf{Symmetric Extremal Dependence Index (SEDI)} is a skill score designed for rare event verification, as it is formulated to be independent of the event's base rate:
\begin{equation}
  \text{SEDI} = \frac{\ln F - \ln H + \ln(1 - H) - \ln(1 - F)}{\ln F + \ln H + \ln(1 - H) + \ln(1 - F)},
  \label{eq:sedi}
\end{equation}
where $H$ is the Hit Rate (POD) and $F$ is the False Alarm Rate ($FP/(FP+TN)$). Its logarithmic structure ensures stability for low event counts, making it robust for extreme event evaluation. SEDI range is typically [-1, 1].

\textbf{Fractions Skill Score (FSS)} is a spatial metric that compares the fractional coverage of an event in a forecast to observations within a neighborhood window. It is tolerant of small location errors:
\begin{equation}
    \text{FSS}(n) = 1 - \frac{\text{MSE}(n)}{\text{MSE}_{ref}(n)}.
    \label{eq:fss}
\end{equation}
FSS is derived from the Mean Squared Error (MSE) of the fractions, normalized by a reference MSE. FSS is used to assess forecast skill across various spatial scales by adjusting the neighborhood size $n$. Its range is $[0, 1]$.

\subsubsection*{Feature quality metrics}\label{subseca52}
We employed a range of metrics to quantify and evaluate the quality of the extracted features. The fundamental principle is to assess feature separability: high-quality features are those where different rainfall classes form distinct and compact clusters in the feature space, making them easily distinguishable for a classifier. A model that learns highly separable features is better equipped to discern between different rainfall intensities, especially rare heavy rainfall events. To address the severe class imbalance and computational challenges inherent in pixel-level analysis, all metrics are computed on a representative subset of the data. This subset is generated using a progressive sampling strategy where the number of samples drawn from each class is adjusted based on its original frequency. This process yields a more balanced feature distribution, allowing for a more robust evaluation of the model's ability to represent critical, rare events. The following metrics are then computed on this progressively sampled set:

\textbf{Intra-Category Similarity (IntraSim)} measures the cohesion within categories by quantifying how similar data points are within the same category using cosine similarity. It is a key indicator for evaluating feature extraction quality:

\begin{equation}
\text{IntraSim} = \frac{1}{|L|} \sum_{l \in L} \frac{1}{|C_l|(|C_l|-1)} \sum_{\substack{i,j \in C_l \\ i \neq j}} \cos(x_i, x_j)
\end{equation}

where $\cos(x_i, x_j) = \frac{x_i \cdot x_j}{\|x_i\| \cdot \|x_j\|}$ is the cosine similarity between normalized feature vectors $x_i$ and $x_j$, $C_l$ is the set of samples in category $l$, and $|C_l|$ is the number of samples in that category. The constraint $i \neq j$ excludes self-similarity. This score represents the average cosine similarity between all sample pairs within each category. A higher value indicates greater intra-category similarity and thus better feature discriminability.

\textbf{Similarity Margin (SimMargin)} measures the separability between different categories by measuring the margin between intra-category and inter-category similarities in the featur e space:

\begin{equation}
\text{SimMargin} = \text{IntraSim} - \text{InterSim}
\end{equation}

where $\text{IntraSim}$ is the average cosine similarity within categories and $\text{InterSim}$ is the average cosine similarity between different categories:

\begin{equation}
\text{InterSim} = \frac{1}{|L|(|L|-1)/2} \sum_{\substack{l_1, l_2 \in L \\ l_1 \neq l_2}} \frac{1}{|C_{l_1}| \cdot |C_{l_2}|} \sum_{i \in C_{l_1}} \sum_{j \in C_{l_2}} \cos(x_i, x_j)
\end{equation}

where $C_{l_1}$ and $C_{l_2}$ represent samples in categories $l_1$ and $l_2$ respectively. SimMargin represents the difference between how similar samples are within their own categories versus how similar they are to samples from other categories. A higher value indicates better feature discriminability with tight intra-category clustering and well-separated inter-category boundaries.

\textbf{K-Nearest Neighbor Accuracy (KNNAcc)} measures feature quality by measuring the classification accuracy of a 1-nearest neighbor (1-NN) classifier based on cosine similarity with a leave-one-out evaluation. For each sample, the most similar sample (excluding itself) is selected as the neighbor, and its label is used as the prediction. This provides a direct assessment of the discriminative power of the features:
\begin{equation}
\text{KNNAcc} = \frac{1}{N} \sum_{i=1}^{N} \mathbb{I}(y_i = \hat{y_i}),
\end{equation}
where $y_i$ is the true label, $\hat{y_i}$ is the label predicted by the 1-NN classifier using cosine similarity, and $\mathbb{I}$ is the indicator function. The score is the fraction of correct predictions, with higher values indicating more useful features for classification.

\textbf{Fisher Score (Fisher)} measures feature discriminability by measuring the ratio of between-class scatter to within-class scatter. It is computed as the trace of the between-class scatter matrix divided by the trace of the within-class scatter matrix:
\begin{equation}
  \text{Fisher} = \frac{\text{trace}(S_b)}{\text{trace}(S_w)},
\end{equation}
where $S_b$ is the between-class scatter matrix defined by the deviation of each class mean from the overall mean, and $S_w$ is the within-class scatter matrix defined by the deviation of samples from their corresponding class means. A higher score indicates greater separation among classes relative to their internal variance, thus reflecting stronger discriminative power of the features.

\textbf{Calinski–Harabasz Index (CH)} measures the quality of features by calculating the ratio of between-category dispersion to within-category dispersion. It reflects how well the categories are separated while maintaining compactness within each category:
\begin{equation}
  \text{CH} = \frac{\text{tr}(B_c)/(c-1)}{\text{tr}(W_c)/(N-c)},
\end{equation}
where $B_c$ is the between-category scatter matrix, $W_c$ is the within-category scatter matrix, $c$ is the number of categories, and $N$ is the total number of samples. A higher CH value indicates more distinct and better-separated categories with greater compactness inside each category.

\textbf{SVM F1 score (F1$_{\text{SVM}}$)} evaluates the overall quality of the features in a direct supervised learning context. A Support Vector Machine (SVM) classifier is trained on the extracted features, and its performance is subsequently measured using the F1 score on a held-out test set. Since the F1 score is the harmonic mean of precision and recall, it provides a balanced assessment of the features' ability to support a discriminative model. A higher F1$_{\text{SVM}}$ score indicates that the features are more effective for building a successful classifier.

\section{Uncertainty quantification}\label{seca6}
To quantify pixel-wise predictive uncertainty of the DPSformer, we employ Mondrian-Conformal Classification (MCC) to construct $(1-\alpha)$ conformal prediction sets for each pixel, with $\alpha=0.05$ for $95\%$ coverage. Conformal prediction provides post-hoc statistical calibration with guaranteed marginal coverage, without re-training or resampling the base model.

Let $f_{\theta}(x) \in \mathbb{R}^C$ denote the raw logits output of the trained segmentation network for an input image $x$, where $C$ is the number of classes. The softmax probability for class $c$ is given by:

\begin{equation}
p_c(x) = \frac{\exp(f_{\theta}(x)_c)}{\sum_{j=1}^{C} \exp(f_{\theta}(x)_j)}.
\label{eq:softmax}
\end{equation}

For a pixel with ground-truth label $y$, we define the \emph{conformity score} as:
\begin{equation}
s(x, y) = 1 - p_y(x).
\label{eq:score}
\end{equation}

In the Mondrian variant, we compute class-conditional thresholds $q_k$ from a calibration set. For class $k$, the empirical $(1-\alpha)$-quantile is obtained conservatively as:
\begin{equation}
q_k = s_{(\lceil (1-\alpha)(N_k + 1) \rceil)},
\label{eq:quantile}
\end{equation}
where $N_k$ is the number of calibration pixels in class $k$ and $s_{(m)}$ denotes the $m$-th order statistic of the conformity scores in that class.

At test time, the conformal prediction set for a pixel $x_i$ is:

\begin{equation}
\mathcal{S}(x_i) = \{\, k \in \{1,\dots,C\} \mid 1 - p_k(x_i) \le q_k \,\}.
\label{eq:predset}
\end{equation}

We assess the conformal predictor on an independent test set using two metrics:

\textbf{Prediction Interval Coverage Probability (PICP)} measures the proportion of pixels whose true label lies within the prediction set:

\begin{equation}
\mathrm{PICP} = \frac{1}{N} \sum_{i=1}^{N} \mathbf{1}\big( y_i \in \mathcal{S}(x_i) \big),
\label{eq:picp}
\end{equation}
where $\mathbf{1}(\cdot)$ denotes the indicator function.

\textbf{Average Set Size} quantifies the average number of candidate classes per pixel, reflecting the sharpness of uncertainty estimates:

\begin{equation}
\overline{|\mathcal{S}|} = \frac{1}{N} \sum_{i=1}^{N} |\mathcal{S}(x_i)|.
\label{eq:avgset}
\end{equation}

A PICP close to $1-\alpha$ indicates well-calibrated uncertainty, while a small $\overline{|\mathcal{S}|}$ reflects high discriminative precision.

\section{More implementation details}\label{seca7}
This study uses two types of public datasets: TIGGE forecast data and TRMM 3B42 V7 satellite rainfall products. The former is obtained through the ECMWF API, and the latter is retrieved from the NASA GES DISC portal. Both are open licenses, and the data acquisition scripts and logs are made public with the code base. Data preprocessing first performs time alignment. TIGGE daily 6-hour interval forecasts starting at 00:00 UTC and 12:00 UTC are selected (6, 12, 18, and 24 hours), and their cumulative precipitation is calculated by 6-hour difference. All variables are z-score standardized based on the mean and standard deviation of the training set. In order to obtain more samples, this study does not construct different correction models for different forecast times. In other words, the atmospheric parameters at different forecast times are used to train the correction model, and the obtained correction model is evenly applied to different forecast times. To avoid information leakage, the data are divided into a training set (11,688 samples) for 2007–2010, a validation set (2,920 samples) for 2011, and a test set (2,928 samples) for 2012. The model input includes 27 TIGGE variables (see Table \ref{tab:meteorological variables}), which come from the control members. The experiment was performed on an NVIDIA RTX A6000 GPU using PyTorch and CUDA. DPSformer uses SegFormer-b0 as the backbone network and adopts AdamW as the optimizer. The specific training parameters and hyperparameters are shown in Table \ref{tab:all_hyperparams}.

\begin{table}[htbp]
\centering
\caption{\textbf{Meteorological variables from the TIGGE dataset used as inputs for the correction models of rainfall forecasts.} These variables encompass key atmospheric parameters at various pressure levels and surface conditions.}
\label{tab:meteorological variables}%
\begin{tabular}{lll}
\toprule
Meteorological variables & Levels & Units \\
\midrule
Temperature & 500/700/850/925 hPa & K \\
Geopotential height & 500/700/850/925 hPa & gpm \\
U component of wind & 500/700/850/925 hPa & m s$^{-1}$ \\
V component of wind & 500/700/850/925 hPa & m s$^{-1}$ \\
Specific humidity & 500/700/850/925 hPa & kg kg$^{-1}$ \\
2 m temperature & SURFACE & K \\
10 m u component of wind & SURFACE & m s$^{-1}$ \\
10 m v component of wind & SURFACE & m s$^{-1}$ \\
Total precipitation  & SURFACE & kg m$^{-2}$ \\
Total column water & SURFACE & kg m$^{-2}$ \\
Convective available potential energy & SURFACE & J kg$^{-1}$ \\
Mean sea level pressure & SURFACE & Pa \\
\bottomrule
\end{tabular}
\end{table}   

\begin{table}[!t]
\centering
\makeatletter
\setlength{\@fptop}{0pt}    
\makeatother
\caption{\textbf{Training and loss function hyperparameters for the DPSformer model.} These hyperparameters were tuned based on the validation set to balance model convergence and generalization in the rainfall correction task. The values include optimizer settings and weights for the dual loss components.}
\label{tab:all_hyperparams}
\begin{tabular}{lclc}
\toprule
\textbf{Training Hyperparameter} & \textbf{Value} & \textbf{Loss Hyperparameter} & \textbf{Value} \\
\midrule
Learning rate           & 1e-3        & Spatial Loss Weight $\gamma$ & 0.7 \\
Weight decay            & 1e-4        & Dice Loss Weight $\alpha $  & 0.5 \\
Batch size              & 64          & LA $\tau$   & 0.5 \\
Epochs                  & 30          & BLV $\sigma$ & 0.5 \\
\bottomrule
\end{tabular}
\end{table}



\end{document}